\documentclass[10pt,letterpaper]{article}
\usepackage{lmodern}
\usepackage{alibaba-report}
\usepackage{colortbl,placeins,afterpage}

\usepackage{amsmath,amsfonts,bm}

\def\eqref#1{equation~\ref{#1}}

\def\1{\bm{1}}

\DeclareMathAlphabet{\mathsfit}{\encodingdefault}{\sfdefault}{m}{sl}
\SetMathAlphabet{\mathsfit}{bold}{\encodingdefault}{\sfdefault}{bx}{n}

\definecolor{dreamrealmshade}{RGB}{252,244,224}
\definecolor{cvprblue}{rgb}{0.21,0.49,0.74}
\DeclareTextFontCommand{\yuviontablebold}{\bfseries}
\AtBeginEnvironment{table}{\let\textbf\yuviontablebold}
\AtBeginEnvironment{table*}{\let\textbf\yuviontablebold}
\AtBeginEnvironment{algorithm}{\let\textbf\yuviontablebold}
\newcommand{\resultdelta}[2]{\makebox[2.5em][r]{#1}\kern.12em\rlap{\textsubscript{#2}}\phantom{\textsubscript{+00.00}}}
\newcommand{\shortpapertitle}{DreamRealm: Exploring and Materializing Hallucination Worlds for Trustworthy Vision-Language Models}
\title{DreamRealm: Exploring and Materializing Hallucination Worlds for Trustworthy Vision-Language Models}
\author{%
  Benlei Cui$^{*}$\quad Chunyang Chai$^{*}$\quad Yiliang Zhang$^{*\dagger}$\quad Xiaoqian Xia\\
  Longtao Huang\quad Yuwen Zhai\quad Song Li\quad Chengwen Yao\\
  Hui Xue\quad Jingqun Tang\quad Haiwen Hong$^{\dagger}$\\[3pt]
  {\reportaffiliationfont Yuvion Team, Alibaba Group}\\
  {\normalfont\fontsize{9}{12}\selectfont\href{mailto:cuibenlei.cbl@alibaba-inc.com}{\texttt{cuibenlei.cbl@alibaba-inc.com}}}\\
  {\normalfont\fontsize{8.5}{11}\selectfont $^{*}$Equal contribution.\quad $^{\dagger}$Corresponding authors.}}
\renewcommand{\authorcontributions}{%
  \textbf{Benlei Cui}: Conceptualization, project design, lead writing, and principal experiments;
  \textbf{Chunyang Chai}: Experiments;
  \textbf{Yiliang Zhang}: Comparative-method experiments and evaluation;
  \textbf{Xiaoqian Xia}: Data-processing pipeline.}
\AtBeginDocument{\hypersetup{pdfauthor={Benlei Cui, Chunyang Chai, Yiliang Zhang, Xiaoqian Xia, Longtao Huang, Yuwen Zhai, Song Li, Chengwen Yao, Hui Xue, Jingqun Tang, Haiwen Hong}}}
\newcommand{\dreamrealmresources}{%
\par\vspace{3pt}\begingroup
\renewcommand{\paperlinkglyph}[1]{%
  \ifstrequal{##1}{GitHub}{\includegraphics[height=16pt]{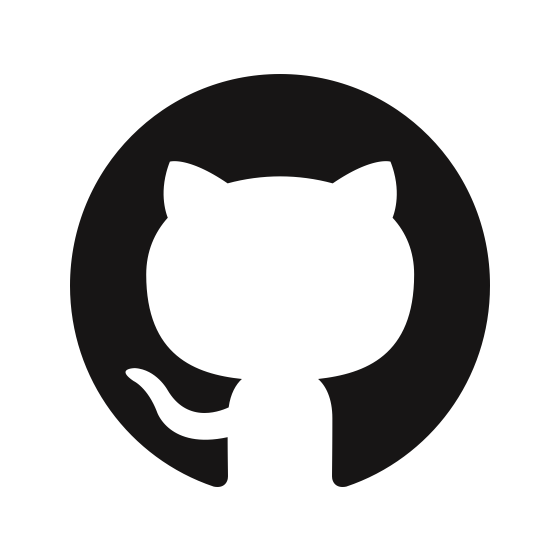}}{%
  \ifstrequal{##1}{Hugging Face}{\includegraphics[height=14pt]{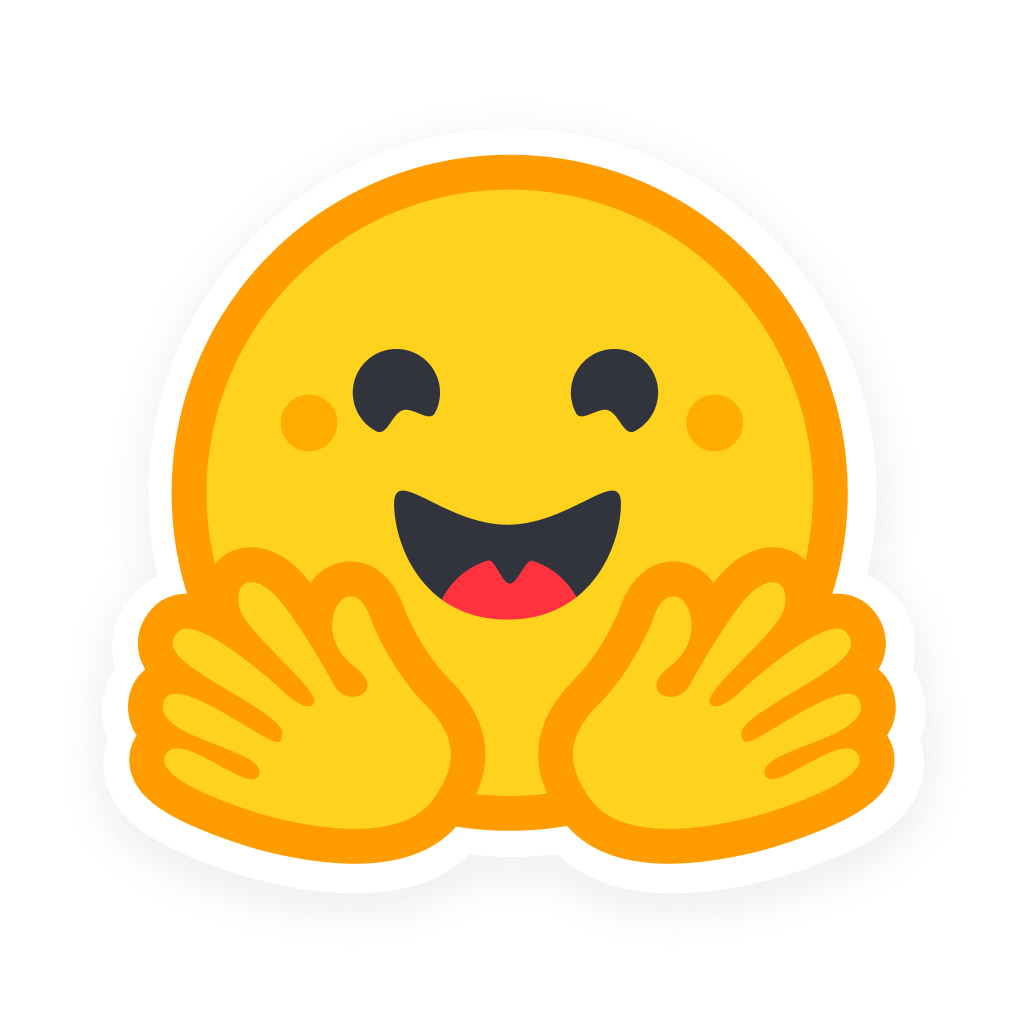}}{}}}
\renewcommand{\paperlinkentry}[2]{\href{##1}{%
  \begin{tikzpicture}[baseline=(label.base)]
  \node[minimum width=112pt,minimum height=16pt,inner sep=0pt] (entry) {};
  \node[anchor=west,inner sep=0pt] at (entry.west) {\paperlinkglyph{##2}};
  \node[anchor=west,inner sep=0pt,text=reportink] (label)
    at ([xshift=22pt]entry.west) {\MakeUppercase{##2}};
  \draw[reportaccent,line width=.6pt,line cap=round,line join=round]
    ([xshift=-5pt,yshift=-1.5pt]entry.east)--([yshift=3pt]entry.east)
    ([xshift=-3.5pt,yshift=3pt]entry.east)--([yshift=3pt]entry.east)--([yshift=-.5pt]entry.east);
  \draw[reportrule,line width=.4pt]
    ([yshift=-4.5pt]label.base -| entry.west)--([yshift=-4.5pt]label.base -| entry.east);
  \draw[reportaccent,line width=.65pt]
    ([yshift=-4.5pt]label.base -| entry.west)--++(8pt,0);
  \end{tikzpicture}}}
\sffamily\fontsize{8.5}{11}\selectfont
\noindent\makebox[\linewidth][c]{%
  \paperlinkentry{https://github.com/Alibaba-VELLDEPTH/DreamRealm}{GitHub}\hspace{32pt}%
  \paperlinkentry{https://huggingface.co/Alibaba-VELLDEPTH/Qwen35-9b-DreamRealm}{Hugging Face}}
\par\endgroup\vspace{7pt}
}
\AtEndEnvironment{abstract}{\dreamrealmresources}

\begin{document}
\maketitle
\begin{abstract}
Multimodal large language models (MLLMs) remain prone to visual hallucinations when correct responses depend on subtle visual details. One line of post-training methods constructs visual contrasts through image retrieval, generation, or editing to improve fine-grained perception. However, these contrasts are not built around the model's own hallucinations, making it difficult to ensure that supervision targets its actual perceptual confusions. Another line suppresses hallucinated outputs through response correction or preference learning under fixed image conditions. How humans correct misconceptions offers a way to extend this supervision. Consider a child who mistakes medicine tablets for candy or cookies: simply saying ``This is not candy'' is not enough. A parent also shows the child what real candy and cookies look like. In effect, this assigns each mistaken judgment to a visual input for which it would be correct. This motivates us to complement hallucination correction by teaching models to assign hallucinated responses to visual conditions that make them valid. We introduce \textbf{DreamRealm}, a post-training framework that explores diverse latent hallucinated responses, minimally edits original images to make them valid, and trains the model to assign each response to its supporting image. Our data construction pipeline, \textbf{RealmForgePipe}, mines observed and latent hallucinations from a target model and materializes them through image editing. Using hallucinations mined separately from each target model studied in this work, we construct the Realized Hallucination Dataset (\textbf{RealmSet}). Our training objective, \textbf{RealmMatch}, constructs a likelihood score matrix across image--response combinations and combines supervised fine-tuning on matched image--response pairs with bidirectional contrastive losses to reinforce correct matches and suppress mismatches. Across three target models and six visual perception and hallucination benchmarks, DreamRealm improves performance in 16 of 18 model--benchmark settings. On Qwen3.5-9B, DreamRealm improves aAcc on the hallucination benchmark HallusionBench by 8.4 percentage points. We will release code, weights, and RealmSet upon publication. 
\end{abstract}
\begin{figure}[t]
  \centering
  \includegraphics[width=\linewidth]{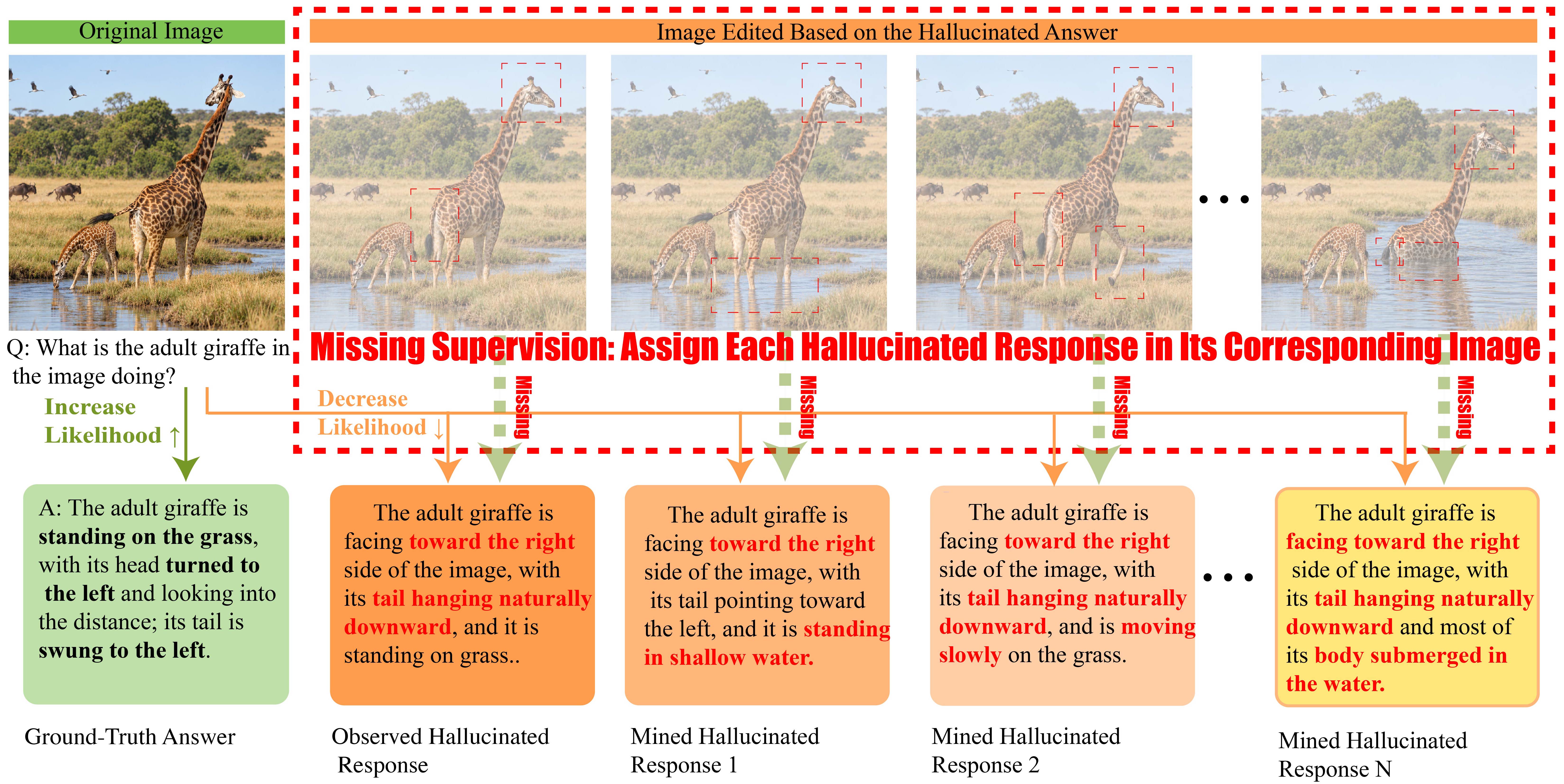}
  \caption{Motivation of DreamRealm. Existing post-training methods correct hallucinated responses, while supervision on the visual conditions that would make them valid remains underexplored. Inspired by how humans correct misconceptions, DreamRealm materializes these conditions and teaches the model to match hallucinated responses to their supporting images.}\label{fig:intro-motivation}
\end{figure}
\section{Introduction}
\label{sec:intro}

Recent multimodal large language models (MLLMs) have demonstrated strong capabilities in visual understanding~\citep{scholar_qwen25vl,scholar_internvl35}. However, these models still generate content that contradicts image facts or lacks support from visual evidence, a phenomenon known as visual hallucination~\citep{scholar_pope,scholar_hallusionbench,scholar_amber}. This problem is particularly pronounced when correct responses depend on subtle visual differences~\citep{scholar_mmvp,scholar_hallusionbench}. Such discrepancies between model outputs and visual evidence remain a major obstacle to the reliable application of multimodal models.

A line of post-training work mitigates visual hallucinations through response correction, contrastive supervision, and preference learning~\citep{scholar_rlhfv,scholar_rlaifv,scholar_halva}. These methods treat images as a fixed condition and teach models to reject or revise hallucinated responses. Another line of work constructs visual contrasts through image generation or retrieval, helping models understand the relationship between fine-grained visual changes and different responses~\citep{scholar_symmpo,scholar_icvco,scholar_med}. However, these visual contrasts are not constructed specifically around the target model's hallucinations, making it difficult to ensure that supervision focuses on its actual perceptual confusions.

How humans correct misconceptions offers an intuitive starting point for designing supervision around specific perceptual confusions. When a child mistakes a tablet for candy or a cookie, a parent may point out the mistake and show the child what real candy and cookies look like. In effect, this correction assigns each mistaken judgment to a concrete visual example for which it would be correct. As illustrated in Figure~\ref{fig:intro-motivation}, existing post-training methods for vision-language models suppress a range of hallucinated responses but lack comparable supervision for hallucination assignment. Inspired by this process, we mine diverse latent hallucinated responses from the model, construct visual conditions under which each response is correct, and train the model to match each response to its supporting image.

To this end, we introduce \textit{DreamRealm}, a post-training framework centered on \emph{matching hallucinated responses to their supporting images}. We first expose different error patterns through repeated high-temperature sampling from the target model, then expand the set of latent hallucinated responses by exploiting the enumerable key attributes of multimodal perception questions. We select hallucinated responses that the model is more inclined to produce based on their likelihood scores. An image-editing pipeline then minimally edits the original images to make the selected responses valid, materializing the model's hallucination world. Using this pipeline, we construct the \emph{Realized Hallucination Dataset} (\textbf{RealmSet}) from hallucinations mined separately from each target model studied in this work. For training, we introduce RealmMatch, which constructs a likelihood score matrix by cross-pairing the original and materialized images with the ground-truth and selected hallucinated responses. By combining supervised fine-tuning on matched image--response pairs with bidirectional contrastive losses, RealmMatch trains the model to suppress mismatched responses while assigning each hallucinated response to its corresponding supporting image. Across three target models---Qwen3.5-9B~\citep{scholar_qwen35}, Qwen2.5-VL-7B~\citep{scholar_qwen25vl}, and InternVL3.5-8B~\citep{scholar_internvl35}---and six visual perception and hallucination benchmarks, DreamRealm demonstrates consistent performance gains in 16 of 18 model--benchmark settings. On Qwen3.5-9B, DreamRealm yields gains of 8.4 percentage points in aAcc on the hallucination benchmark HallusionBench and 11.3 percentage points in pair accuracy on the fine-grained visual perception benchmark MMVP. We further evaluate hallucination-guided data construction and compare RealmMatch with alternative training objectives under the same initialization and training data.

Our main contributions are as follows:
\begin{itemize}
    \item \textbf{RealmForgePipe: A data pipeline for exploring and materializing a model's own hallucinations.} It derives image-editing targets from observed and mined latent hallucinations and constructs visual conditions that make these responses valid. Using this pipeline, we build the \emph{Realized Hallucination Dataset} (\textbf{RealmSet}) from hallucinations mined separately from each target model studied in this work.
    \item \textbf{RealmMatch: learning to assign hallucinated responses to their supporting images.} We combine supervised fine-tuning on matched image--response pairs with bidirectional contrastive learning to teach the model to assign each hallucinated response to its supporting image.
    \item \textbf{Empirical validation and ablation analysis.} We evaluate three target models on six visual perception and hallucination benchmarks, with separate ablations for the data pipeline and training method. The results highlight DreamRealm's potential to improve visual perception and mitigate hallucinations.
\end{itemize}

\section{Related Work}
\label{sec:related-work}

\subsection{Hallucination-aware Fine-tuning}
\looseness=1 Several approaches mitigate hallucinations by correcting responses while keeping the image fixed. HALVA~\citep{scholar_halva} applies contrastive supervision to phrases that differ between faithful and hallucinated responses. RLAIF-V~\citep{scholar_rlaifv} learns from preference data derived from model feedback. FINER-Tuning~\citep{scholar_finer} uses fine-grained contradictions to construct negative queries for DPO training. REVERSE~\citep{scholar_reverse} uses uncertainty tokens and retrospective resampling to revise potentially hallucinated content. These methods focus on correcting responses to a given image, leaving supervision through changes to the visual input largely unexplored.

\subsection{Visual Contrast and Counterfactual Learning}
Visual contrasts have been used in both training and inference. SymMPO~\citep{scholar_symmpo} pairs similar images with corresponding responses for preference learning. At inference time, CoCI~\citep{scholar_coci} uses retrieved images as contrastive inputs, while visual contrastive decoding (VCD)~\citep{scholar_vcd} contrasts response distributions under different visual inputs.

\looseness=1 Another line of work constructs visual contrasts by introducing controlled changes through image generation and editing. IC-VCO~\citep{scholar_icvco} draws its editing targets from existing SymMPO preference data and uses semantic editing to construct contrastive images for preference optimization. Within predefined semantic categories, MED~\citep{scholar_med} generates minimal editing instructions based on image content and uses the resulting image pairs and fine-grained difference descriptions for supervised fine-tuning. However, these visual contrasts are not designed specifically around the target model's hallucinations, making it difficult to ensure that the resulting supervision directly addresses its perceptual confusions.

DreamRealm differs from prior work in two ways. First, we mine diverse latent hallucinated responses and train the model to assign each response to the visual conditions that make it correct. Second, we use the target model's own hallucinations to guide image editing, so the resulting visual contrasts address its specific perceptual confusions.

\begingroup
\raggedbottom
\makeatletter
\renewcommand{\paragraph}{\@startsection{paragraph}{4}{\z@}{5pt plus 1pt minus 1pt}{-0.6em}{\normalfont\normalsize\bfseries}}
\makeatother
\setlength{\abovedisplayskip}{6pt plus 1pt minus 1pt}
\setlength{\belowdisplayskip}{6pt plus 1pt minus 1pt}
\setlength{\abovedisplayshortskip}{3pt plus 1pt minus 1pt}
\setlength{\belowdisplayshortskip}{5pt plus 1pt minus 1pt}
\setlength{\dbltextfloatsep}{12pt plus 1pt minus 1pt}
\section{Method}
\label{sec:method}

Figure~\ref{fig:dreamrealm-overview} shows the DreamRealm pipeline. RealmForgePipe collects observed hallucinations, mines latent hallucinations, and ranks candidate responses by likelihood (Section~\ref{sec:hallucination-mining}). It constructs images that support the selected responses through minimal editing and verification (Section~\ref{sec:hallucination-materialization}). Section~\ref{sec:realmset} describes the resulting \emph{Realized Hallucination Dataset} (\textbf{RealmSet}). Section~\ref{sec:realmmatch} presents RealmMatch, which combines supervised fine-tuning and bidirectional contrastive learning to assign hallucinated responses to supporting images while retaining error-correction supervision on the original images.

\begin{figure}[!htb]
  \centering
  \includegraphics[width=\textwidth]{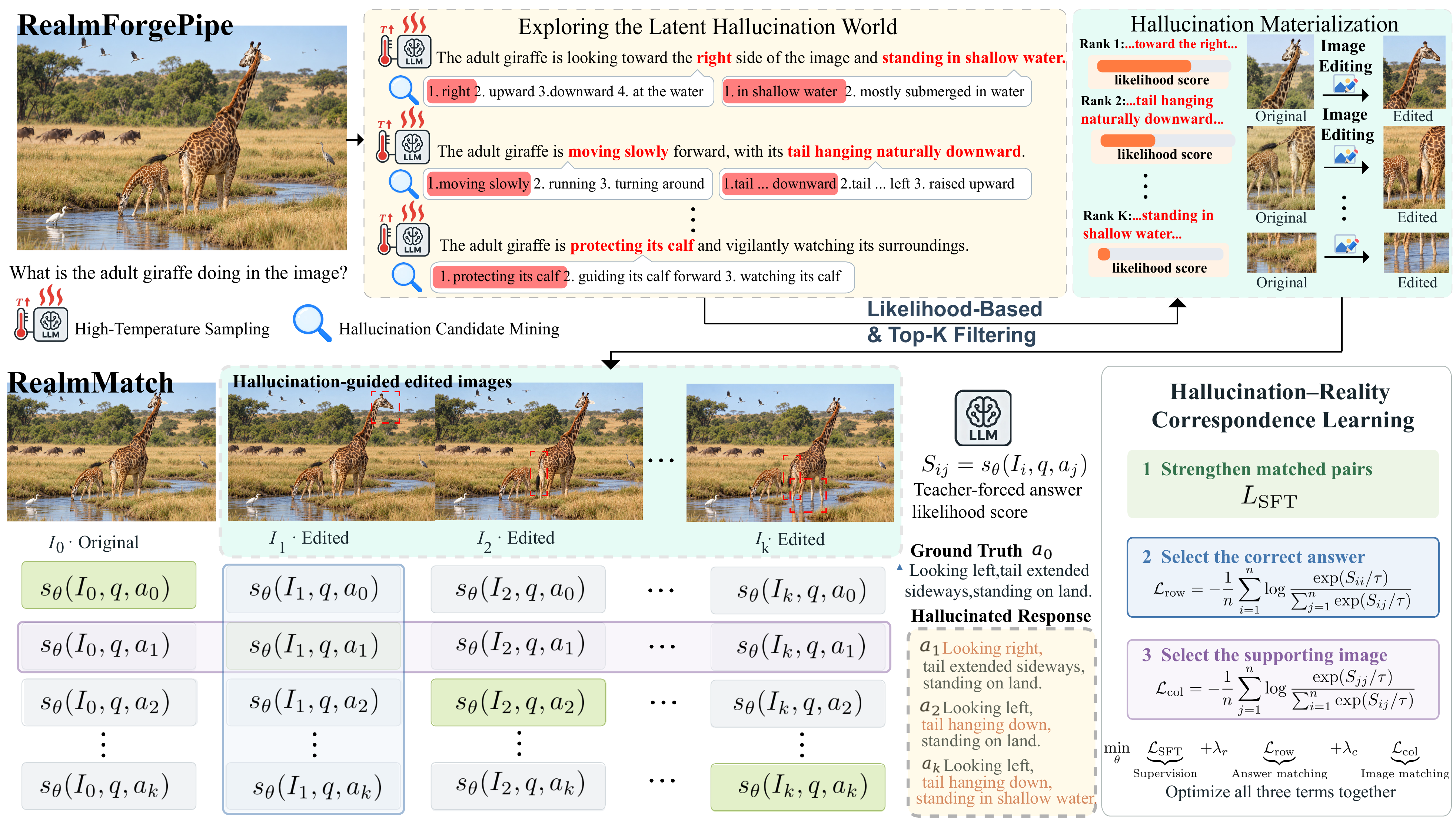}
  \caption{Overview of DreamRealm. \textbf{Top:} RealmForgePipe mines hallucinated responses through high-temperature sampling and key-attribute enumeration, selects candidates through likelihood ranking and Top-K filtering, and edits images to make the selected hallucinated responses valid. \textbf{Bottom:} RealmMatch constructs an image--response score matrix from original images with their ground-truth answers and edited images with their corresponding hallucinated responses. Through SFT on matched image--response pairs and bidirectional contrastive learning, it assigns hallucinated responses to their supporting images, learning the correspondence between responses and visual conditions.}
  \label{fig:dreamrealm-overview}
\end{figure}

\subsection{RealmForgePipe: Mining Hallucinations}
\label{sec:hallucination-mining}

\paragraph{Problem formulation.}
Let $\mathcal{D}=\{(I_i,q_i,y_i^\star)\}_{i=1}^{N}$ denote a collection of seed examples, where $I_i$ is an image, $q_i$ is a perception question, and $y_i^\star$ is the ground-truth answer. For a target MLLM $M_\theta$, our goal is to identify a set $\mathcal{E}_i$ of incorrect responses that receive high likelihood under the model given $(I_i,q_i)$. We refer to incorrect responses produced under greedy decoding as \emph{observed hallucinations}. To explore \emph{latent perceptual hallucinations}, we use high-temperature sampling and key-attribute enumeration to uncover a broader set of plausible incorrect responses. We mine latent perceptual hallucinations only for inputs on which the target model's greedy response is incorrect.

\begin{samepage}
\paragraph{Diversified candidate elicitation.}
We first use high-temperature sampling to explore diverse patterns of the model's hallucinated responses across perceptual dimensions. Given $(I_i,q_i)$, we sample $R$ responses at temperature $\tau_s$ to form
\[
\mathcal{S}_i =
\left\{
\widetilde{y}_i^{(r)}
\sim p_\theta(\,\cdot\mid I_i,q_i;\tau_s)
\right\}_{r=1}^{R}.
\]
\end{samepage}
We expand the sampled responses by enumerating alternative values of the attributes queried by $q_i$, forming $\mathcal{H}(q_i)$. In Figure~\ref{fig:dreamrealm-overview}, one sampled response describes the adult giraffe as looking right. Varying its gaze direction yields additional candidates, such as looking upward or downward.

We perform semantic deduplication over all proposed candidates, retaining only one representative from each group of semantically equivalent hallucinated responses to form the candidate pool $\mathcal{C}_i$.

\paragraph{Likelihood-based hallucination ranking.}
We score each candidate $e\in\mathcal{C}_i$ using teacher forcing. All candidates use the same answer template. If the rendered candidate contains tokens $z_{1:T_e}^{e}$, we define its log-likelihood score as the average token log-likelihood over the answer tokens:
\[
s_i(e)=s_\theta(I_i,q_i,e)=
\frac{1}{T_e}
\sum_{t=1}^{T_e}
\log p_\theta
\left(
z_t^{e}\mid I_i,q_i,z_{<t}^{e}
\right).
\]
Here $s_\theta(I,q,a)$ denotes the same mean token log-likelihood for any image $I$, question $q$, and response $a$; thus $s_i(e)=s_\theta(I_i,q_i,e)$.
We apply softmax normalization to the candidate scores to obtain the candidate-relative weights $\rho_i(e)$ within the candidate set:
\[
\rho_i(e)=
\frac{\exp(s_i(e))}
{\sum_{e'\in\mathcal{C}_i}\exp(s_i(e'))}.
\]
Importantly, $\rho_i(e)$ is not a calibrated absolute hallucination probability and depends on the coverage and granularity of $\mathcal{C}_i$.

We then rank the verified errors by $s_i(e)$ and retain the top $K_{\mathrm{mine}}$ candidates, or all remaining candidates when fewer than $K_{\mathrm{mine}}$ survive filtering:
\[
\mathcal{E}_i =
\operatorname{Top}_{K_{\mathrm{mine}};\,e\in\mathcal{C}_i}\; s_i(e).
\]
Figures~\ref{fig:hallucination-mining-examples} and~\ref{fig:hallucination-mining-additional} in Appendix~\ref{app:mining-examples} illustrate mined responses and their candidate-relative weights across action understanding, spatial relations, counting, clock reading, color recognition, and chart analysis. The selected set $\mathcal{E}_i$ is passed to the visual materialization stage, which constructs a hallucination-guided edited image supporting each hallucinated response.

\subsection{RealmForgePipe: Materializing Hallucinations}
\label{sec:hallucination-materialization}

For each selected hallucinated response $y_m\in\mathcal{E}_i$, RealmForgePipe edits the original image $I$ so that $y_m$ correctly answers the same question $q$ on the edited image $I^\star$. A multimodal verifier $\mathcal{V}$ instructs an image generator to change only the visual content needed to support $y_m$ while preserving the rest of the scene. In Figure~\ref{fig:dreamrealm-overview}, the adult giraffe in $I_0$ stands on land looking left, its tail extended sideways. Turning only its head makes the response ``looking right, tail extended sideways, standing on land'' valid in $I_1$. Each edit must pass checks for edit success and visual plausibility and score at least $4/5$ for minimal editing. The verifier screens each edit before human review, as detailed in Appendix~\ref{app:materialization}. Appendix~\ref{app:editing-examples} shows the resulting edits, and Appendix~\ref{app:editing-walkthrough} walks through one example.

Edited image--response pairs that pass per-edit verification proceed to human review. In the first round, annotators check that each edited image supports its assigned response. In the second round, they review every family $\mathcal{G}=(q,\{(I_i,a_i)\}_{i=0}^{K})$ as a whole, where $(I_0,a_0)$ is the original pair and the remaining pairs are edited. They check that each image supports its assigned response and that no other response in the family is also valid for that image. They remove pairs with cross-matches. The acceptance criterion for the retained pairs is
\begin{equation}
\operatorname{Valid}(I_i,q,a_j)=\mathbf{1}[i=j],
\qquad \forall\,i,j\in\{0,\ldots,K\}.
\label{eq:family-exclusivity}
\end{equation}
Only human-approved edits are retained for training; in bidirectional contrastive learning, matched pairs serve as positives and off-diagonal combinations as negatives. The per-edit generation, validation, and retry procedure is provided in Algorithm~\ref{alg:materialization} in Appendix~\ref{app:materialization}, with the human review protocol in Appendix~\ref{app:human-review}.

\subsection{RealmSet: The Realized Hallucination Dataset}
\label{sec:realmset}

We construct the \emph{Realized Hallucination Dataset} (\textbf{RealmSet}) from VC-STaR's VisCoR-55K~\citep{scholar_vcstar} using RealmForgePipe. We mine hallucinated responses separately from three target models---Qwen3.5-9B, Qwen2.5-VL-7B, and InternVL3.5-8B---and use each model's own hallucinations to guide image editing. Each family groups the original image and its ground-truth answer with edited image--response pairs for the same question. A hallucinated response to the original image becomes a correct answer on its corresponding edited image. RealmSet contains 43,483 human-approved edited image--response pairs spanning 22 types of visual perception tasks, including counting, scene description, spatial perception, text recognition, and chart interpretation. Appendix~\ref{app:data-construction} provides the construction details and the full breakdown by question type.

\subsection{RealmMatch: Hallucination Assignment}
\label{sec:realmmatch}

RealmMatch combines supervised learning with bidirectional contrast over image--response likelihoods, learning the supporting image for each response while retaining error correction under the original image.

\paragraph{Image--response likelihood matrix.}
For each retained family $\mathcal{G}$ with $K\geq1$ edits, let $n=K+1$ and $i,j=0,\ldots,K$. Figure~\ref{fig:dreamrealm-overview} displays responses in rows and images in columns. The entry in row $j$ and column $i$ is $T_{ji}=s_\theta(I_i,q,a_j)$, the mean token log-likelihood defined in Section~\ref{sec:hallucination-mining}. In the figure's score notation, $S_{ij}=T_{ji}$. Here $a_0=y^\star$ and $a_i$ for $i>0$ is the hallucinated response used to construct $I_i$. We backpropagate through all scores to update $\theta$.

\paragraph{Bidirectional correspondence learning.}
The loss subscripts $\mathrm{row}$ and $\mathrm{col}$ follow the indexing of $S=T^\top$. To select the correct response for image $I_i$, we fix column $i$ of $T$ and normalize over all responses in the family:
\begin{equation}
\mathcal{L}_{\mathrm{row}}^{\mathrm{edit}}
=-\frac{1}{n}\sum_{i=0}^{K}
\log\frac{\exp(T_{ii}/\tau)}{\sum_{j=0}^{K}\exp(T_{ji}/\tau)}.
\label{eq:realm-row}
\end{equation}
To select the image that supports response $a_j$, we fix row $j$ of $T$ and normalize over all images in the family:
\begin{equation}
\mathcal{L}_{\mathrm{col}}^{\mathrm{edit}}
=-\frac{1}{n}\sum_{j=0}^{K}
\log\frac{\exp(T_{jj}/\tau)}{\sum_{i=0}^{K}\exp(T_{ji}/\tau)}.
\label{eq:realm-col}
\end{equation}
The resulting weights are relative to the candidate images in the family and are computed from the conditional likelihood of the same response under each image. Here $\tau>0$ is the contrastive temperature, distinct from the generation temperature $\tau_s$ used during mining.

For a family with $n$ matched image--response pairs, the SFT loss is the mean negative log-likelihood over those pairs,
\begin{equation}
\mathcal{L}_{\mathrm{SFT}}
=-\frac{1}{n}\sum_{i=0}^{n-1}T_{ii}.
\label{eq:realm-sft}
\end{equation}
The joint training objective is
\begin{equation}
\mathcal{L}_{\mathrm{edit}}
=\mathcal{L}_{\mathrm{SFT}}
+\lambda_r\mathcal{L}_{\mathrm{row}}^{\mathrm{edit}}
+\lambda_c\mathcal{L}_{\mathrm{col}}^{\mathrm{edit}},
\label{eq:realm-edit}
\end{equation}
where $\lambda_r,\lambda_c\geq0$. All three terms are optimized jointly.

\paragraph{Original examples without edited images.}
For examples without an accepted edited image, only the image-conditioned response contrast applies. We construct a record $\mathcal{U}=(I_0,q,a_0,\{b_j\}_{j=1}^{M})$, where $b_j$ are offline-verified incorrect responses and responses equivalent to $a_0$ are excluded. Let $s_+=s_\theta(I_0,q,a_0)$ denote the score of the ground-truth answer and $s_j^-=s_\theta(I_0,q,b_j)$ the score of the $j$th verified negative. We combine the SFT loss from Eq.~(\ref{eq:realm-sft}) with a contrastive loss over these responses:
\begin{equation}
\begin{aligned}
\mathcal{L}_{\mathrm{row}}^{\mathrm{plain}}
&= -\log \frac{\exp(s_+/\tau)}
{\exp(s_+/\tau)+\sum_{j=1}^{M}\exp(s_j^-/\tau)},\\[4pt]
\mathcal{L}_{\mathrm{plain}}
&= \mathcal{L}_{\mathrm{SFT}} + \lambda_r\mathcal{L}_{\mathrm{row}}^{\mathrm{plain}}.
\end{aligned}
\label{eq:realm-plain}
\end{equation}
Only one image is available, so we omit the image contrast. We retain $\lambda_r$ unchanged rather than transferring the absent column weight to it. When $M=0$, the contrastive term vanishes and the objective reduces to SFT. When $M=1$, it becomes a binary logistic preference term plus SFT, without a reference-model likelihood ratio.

\paragraph{Overall training objective.}
The overall training objective takes two forms depending on whether accepted edited data are available. With accepted edits, bidirectional contrast assigns hallucinated responses to images that make them valid; otherwise, the response contrast suppresses these responses under the original image. Both cases retain supervised learning on correct image--response pairs. Let $K_g$ denote the number of accepted edits associated with record $g$. We average the corresponding losses over the $B$ records or families in each batch:
\begin{equation}
\mathcal{L}_{\mathrm{batch}}=\frac{1}{B}\sum_{g=1}^{B}\mathcal{L}_g,
\qquad
\mathcal{L}_g=
\begin{cases}
\mathcal{L}_{\mathrm{edit},g}, & K_g\geq1,\\
\mathcal{L}_{\mathrm{plain},g}, & K_g=0.
\end{cases}
\label{eq:realm-batch}
\end{equation}
We first average the losses within each family and then across the batch.
\par
\endgroup

\begingroup
\raggedbottom
\makeatletter
\renewcommand{\paragraph}{\@startsection{paragraph}{4}{\z@}{5pt}{-1em}{\normalfont\normalsize\bfseries}}
\makeatother
\setlength{\floatsep}{12pt plus 2pt minus 2pt}
\setlength{\dblfloatsep}{12pt plus 2pt minus 2pt}
\setlength{\textfloatsep}{12pt plus 2pt minus 2pt}
\setlength{\dbltextfloatsep}{12pt plus 2pt minus 2pt}
\setlength{\intextsep}{12pt plus 2pt minus 2pt}
\section{Experiments}
\label{sec:experiments}

\subsection{Experimental Setup}
\label{sec:experimental-setup}

We train each of the three target models with RealmMatch on RealmSet examples derived from its own hallucinations. Appendices~\ref{app:training-settings} and~\ref{app:evaluation} give the training settings and evaluation protocol, respectively.

\subsection{Main Comparison}
\label{sec:main-results}

\paragraph{Baseline sources.}
We compare DreamRealm with specialized hallucination-mitigation methods, including VCD~\citep{scholar_vcd}, CoCI-CAP~\citep{scholar_coci}, RLAIF-V~\citep{scholar_rlaifv}, SymMPO~\citep{scholar_symmpo}, mDPO~\citep{scholar_mdpo}, VC-STaR~\citep{scholar_vcstar}, MED~\citep{scholar_med} , IC-VCO~\citep{scholar_icvco}, FINER~\citep{scholar_finer}, IRIS~\citep{scholar_iris}, and C$^2$-DPO~\citep{scholar_c2dpo}.
More details are provided in the appendix~\ref{app:compared-methods}.

Table~\ref{tab:main-comparison} compares specialized hallucination-mitigation methods and each backbone with its DreamRealm adaptation across the six benchmarks. Each adaptation uses hallucinations mined from that same backbone.

\begin{table}[!htb]
\centering
\caption{Main results on six visual perception and hallucination benchmarks. The upper section lists specialized hallucination-mitigation methods. The lower section shows each target model before and after DreamRealm training. All methods are evaluated under the same protocol. HallusionBench scores use only image-bearing questions. Subscripts show changes relative to the raw model. Positive values indicate improvement.}
\label{tab:main-comparison}
\begingroup
\color{black}
\fontsize{8.8}{10.4}\selectfont
\setlength{\tabcolsep}{2pt}
\renewcommand{\arraystretch}{1.16}
\resizebox{\textwidth}{!}{%
\begin{tabular}{@{}l*{8}{c}@{}}
  \toprule\noalign{\vskip-1pt}
  & \multicolumn{2}{c}{HallusionBench} & \multicolumn{2}{c}{MMVP} & AMBER & \shortstack{MMHal-\\Bench} & \shortstack{CV-Bench-\\2D} & ObjectHal \\
  \cmidrule(lr){2-3}\cmidrule(lr){4-5}
  \raisebox{6pt}[0pt][0pt]{Model / Method} & aAcc $\uparrow$ & fAcc $\uparrow$ & Item Acc $\uparrow$ & Pair Acc $\uparrow$ & Avg Acc $\uparrow$ & Hall.\ Rate $\downarrow$ & Macro Acc $\uparrow$ & CHAIRs $\downarrow$ \\
  \midrule
  VCD~\citep{scholar_vcd} & \resultdelta{47.21}{} & \resultdelta{17.05}{} & \resultdelta{63.00}{} & \resultdelta{34.00}{} & \resultdelta{77.71}{} & \resultdelta{54.17}{} & \resultdelta{52.42}{} & \resultdelta{44.33}{} \\
  CoCI-CAP~\citep{scholar_coci} & \resultdelta{54.26}{} & \resultdelta{25.14}{} & \resultdelta{75.67}{} & \resultdelta{54.67}{} & \resultdelta{84.05}{} & \resultdelta{36.46}{} & \resultdelta{67.46}{} & \resultdelta{20.67}{} \\
  RLAIF-V~\citep{scholar_rlaifv} & \resultdelta{54.99}{} & \resultdelta{24.57}{} & \resultdelta{57.33}{} & \resultdelta{24.67}{} & \resultdelta{80.59}{} & \resultdelta{39.58}{} & \resultdelta{59.26}{} & \resultdelta{\textbf{8.33}}{} \\
  SymMPO~\citep{scholar_symmpo} & \resultdelta{41.85}{} & \resultdelta{17.05}{} & \resultdelta{61.67}{} & \resultdelta{29.33}{} & \resultdelta{79.75}{} & \resultdelta{56.25}{} & \resultdelta{61.37}{} & \resultdelta{15.67}{} \\
  mDPO~\citep{scholar_mdpo} & \resultdelta{53.63}{} & \resultdelta{23.12}{} & \resultdelta{64.67}{} & \resultdelta{31.33}{} & \resultdelta{80.25}{} & \resultdelta{45.83}{} & \resultdelta{53.73}{} & \resultdelta{14.00}{} \\
  VC-STaR~\citep{scholar_vcstar} & \resultdelta{61.83}{} & \resultdelta{37.28}{} & \resultdelta{76.33}{} & \resultdelta{56.00}{} & \resultdelta{84.20}{} & \resultdelta{44.79}{} & \resultdelta{74.71}{} & \resultdelta{12.33}{} \\
    MED~\citep{scholar_med} & \resultdelta{63.20}{} & \resultdelta{36.71}{} & \resultdelta{79.00}{} & \resultdelta{60.00}{} & \resultdelta{82.02}{} & \resultdelta{31.25}{} & \resultdelta{69.29}{} & \resultdelta{9.00}{} \\
  IC-VCO~\citep{scholar_icvco} & \resultdelta{62.57}{} & \resultdelta{37.57}{} & \resultdelta{75.00}{} & \resultdelta{52.67}{} & \resultdelta{85.88}{} & \resultdelta{39.58}{} & \resultdelta{70.04}{} & \resultdelta{18.00}{} \\
  FINER~\citep{scholar_finer} & \resultdelta{64.98}{} & \resultdelta{41.33}{} & \resultdelta{77.33}{} & \resultdelta{55.33}{} & \resultdelta{85.74}{} & \resultdelta{32.29}{} & \resultdelta{76.29}{} & \resultdelta{19.33}{} \\
  IRIS~\citep{scholar_iris} & \resultdelta{46.79}{} & \resultdelta{18.79}{} & \resultdelta{63.33}{} & \resultdelta{29.33}{} & \resultdelta{77.59}{} & \resultdelta{52.08}{} & \resultdelta{60.94}{} & \resultdelta{40.33}{} \\
  C$^2$-DPO~\citep{scholar_c2dpo} & \resultdelta{45.95}{} & \resultdelta{18.79}{} & \resultdelta{64.33}{} & \resultdelta{32.00}{} & \resultdelta{76.95}{} & \resultdelta{44.79}{} & \resultdelta{60.87}{} & \resultdelta{16.67}{} \\
  \midrule
  Qwen3.5-9B (Raw)~\citep{scholar_qwen35} & \resultdelta{71.61}{} & \resultdelta{54.91}{} & \resultdelta{80.00}{} & \resultdelta{63.33}{} & \resultdelta{88.34}{} & \resultdelta{47.92}{} & \resultdelta{82.16}{} & \resultdelta{30.33}{} \\
  \rowcolor{dreamrealmshade}
  \hspace{1em}\textbf{+ DreamRealm}& \resultdelta{\textbf{80.02}}{\textcolor[rgb]{0.05,0.35,0.26}{+8.41}} & \resultdelta{\textbf{60.98}}{\textcolor[rgb]{0.05,0.35,0.26}{+6.07}} & \resultdelta{\textbf{86.67}}{\textcolor[rgb]{0.05,0.35,0.26}{+6.67}} & \resultdelta{\textbf{74.67}}{\textcolor[rgb]{0.05,0.35,0.26}{+11.34}} & \resultdelta{\textbf{91.01}}{\textcolor[rgb]{0.05,0.35,0.26}{+2.67}} & \resultdelta{\textbf{27.08}}{\textcolor[rgb]{0.05,0.35,0.26}{+20.84}} & \resultdelta{\textbf{82.89}}{\textcolor[rgb]{0.05,0.35,0.26}{+0.73}} & \resultdelta{8.67}{\textcolor[rgb]{0.05,0.35,0.26}{+21.66}}  \\
  \addlinespace[3pt]
  Qwen2.5-VL-7B (Raw)~\citep{scholar_qwen25vl} & \resultdelta{61.30}{} & \resultdelta{38.73}{} & \resultdelta{77.33}{} & \resultdelta{56.00}{} & \resultdelta{85.22}{} & \resultdelta{38.54}{} & \resultdelta{75.37}{} & \resultdelta{13.33}{} \\
  \rowcolor{dreamrealmshade}
  \hspace{1em}\textbf{+ DreamRealm}& \resultdelta{73.08}{\textcolor[rgb]{0.05,0.35,0.26}{+11.78}} & \resultdelta{48.84}{\textcolor[rgb]{0.05,0.35,0.26}{+10.12}} & \resultdelta{78.33}{\textcolor[rgb]{0.05,0.35,0.26}{+1.00}} & \resultdelta{59.33}{\textcolor[rgb]{0.05,0.35,0.26}{+3.33}} & \resultdelta{86.14}{\textcolor[rgb]{0.05,0.35,0.26}{+0.92}} & \resultdelta{30.21}{\textcolor[rgb]{0.05,0.35,0.26}{+8.33}} & \resultdelta{75.94}{\textcolor[rgb]{0.05,0.35,0.26}{+0.57}} & \resultdelta{11.67}{\textcolor[rgb]{0.05,0.35,0.26}{+1.66}}  \\
  \addlinespace[3pt]
  InternVL3.5-8B (Raw)~\citep{scholar_internvl35}& \resultdelta{70.98}{} & \resultdelta{45.38}{} & \resultdelta{75.67}{} & \resultdelta{53.33}{} & \resultdelta{88.79}{} & \resultdelta{34.38}{} & \resultdelta{78.58}{} & \resultdelta{12.00}{}  \\
  \rowcolor{dreamrealmshade}
  \hspace{1em}\textbf{+ DreamRealm}& \resultdelta{74.24}{\textcolor[rgb]{0.05,0.35,0.26}{+3.26}} & \resultdelta{51.16}{\textcolor[rgb]{0.05,0.35,0.26}{+5.78}} & \resultdelta{77.67}{\textcolor[rgb]{0.05,0.35,0.26}{+2.00}} & \resultdelta{56.67}{\textcolor[rgb]{0.05,0.35,0.26}{+3.33}} & \resultdelta{89.04}{\textcolor[rgb]{0.05,0.35,0.26}{+0.26}} & \resultdelta{\textbf{27.08}}{\textcolor[rgb]{0.05,0.35,0.26}{+7.29}} & \resultdelta{77.59}{\textcolor[rgb]{0.75,0.10,0.10}{-0.98}} & \resultdelta{15.33}{\textcolor[rgb]{0.75,0.10,0.10}{-3.33}}  \\
  \bottomrule
  \end{tabular}
}
\endgroup
\end{table}

\begingroup
\rightskip=0pt plus 1.5em\relax
\paragraph{Comparison with existing hallucination-mitigation methods.}
Table~\ref{tab:main-comparison} compares the end-to-end performance of DreamRealm with that of existing hallucination-mitigation methods. With Qwen3.5-9B as its backbone, DreamRealm achieves the best reported result on seven of eight metrics. On ObjectHal, DreamRealm achieves a CHAIRs score of 8.67, comparable to RLAIF-V's 8.33. On the same Qwen2.5-VL-7B backbone, DreamRealm outperforms VC-STaR, MED, and FINER on both HallusionBench accuracy metrics, AMBER average accuracy, and MMHal-Bench hallucination rate. Its HallusionBench fAcc reaches 48.84\%, exceeding the next-best score (41.33\% for FINER) by 7.51 percentage points. Overall, DreamRealm ranks among the leading methods for visual hallucination mitigation in this comparison.
\par\endgroup

\paragraph{DreamRealm improves all three models.}
Relative to their original checkpoints, Qwen3.5-9B and Qwen2.5-VL-7B improve on all eight metrics. InternVL3.5-8B improves on six, with comparable performance on CV-Bench-2D and a decline on ObjectHal. All three models improve on both accuracy metrics on the hallucination benchmark HallusionBench, with Qwen2.5-VL-7B gaining \textbf{11.78} percentage points in aAcc and \textbf{10.12} in fAcc. On MMVP, a benchmark for fine-grained visual perception, Qwen3.5-9B gains \textbf{11.34} percentage points in pair accuracy. Figure~\ref{fig:main-showcase} compares Qwen3.5-9B's responses before and after DreamRealm training, with additional examples in Appendix~\ref{app:qualitative-comparison}.

\begin{figure}[!htb]
\setlength{\abovecaptionskip}{4pt}
\setlength{\belowcaptionskip}{0pt}
\centering
\includegraphics[width=\textwidth]{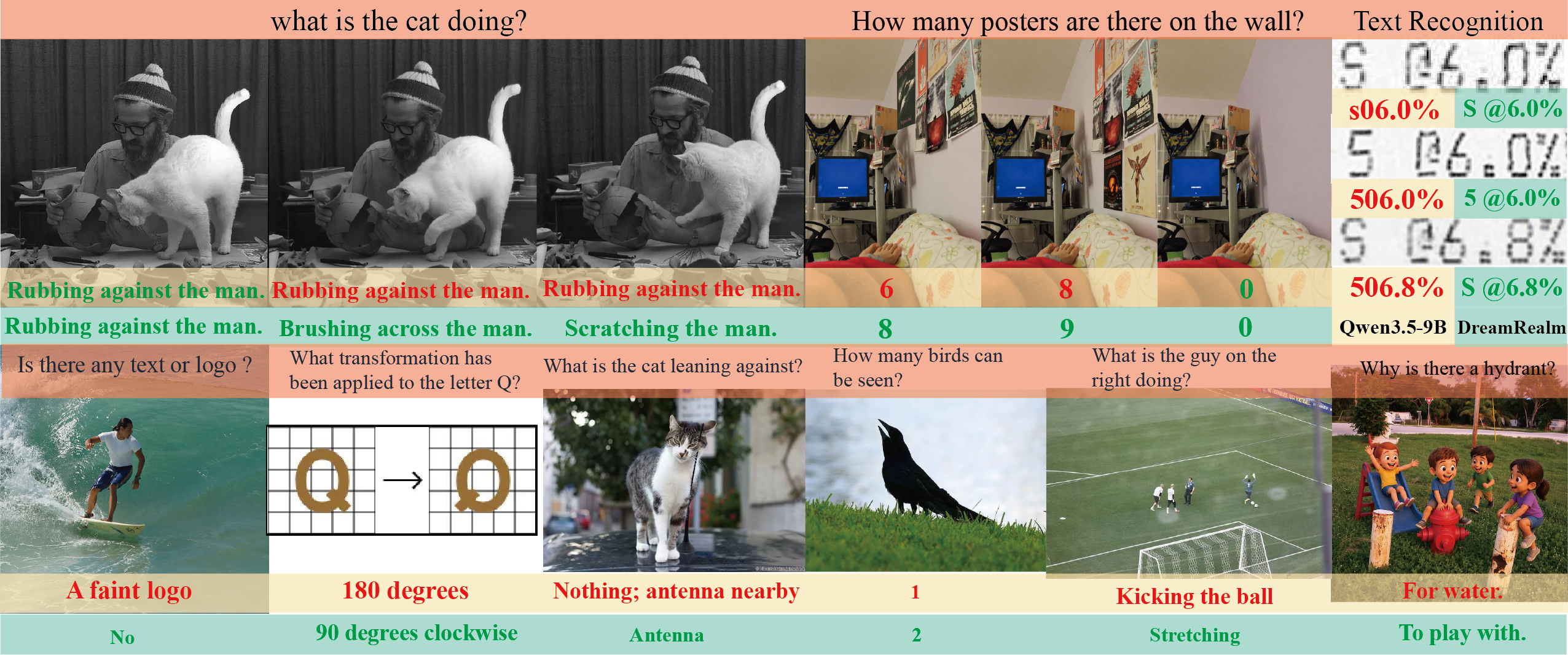}
\caption{Selected examples of Qwen3.5-9B responses before and after DreamRealm training. Yellow/mint bands denote Qwen3.5-9B/DreamRealm; red/green text marks incorrect/correct answers. Responses are condensed. }
\label{fig:main-showcase}
\end{figure}

A breakdown of the benchmark gains shows that all three models improve on tasks well represented in RealmSet. On HallusionBench, chart interpretation accuracy increases by 10.77, 22.31, and 3.85 percentage points for Qwen3.5-9B, Qwen2.5-VL-7B, and InternVL3.5-8B, respectively. Chart interpretation accounts for 8.88\% of the human-approved edited pairs in RealmSet (Appendix~\ref{app:data-categories}). On AMBER's number-verification questions, accuracy improves by 4.54, 2.61, and 1.45 percentage points for the three models, respectively, consistent with the prevalence of counting questions in RealmSet. These gains show that DreamRealm improves performance on these visual tasks and suggest that broadening RealmSet's task coverage could extend its benefits to other tasks.

\subsection{Ablation Study}
\label{sec:ablation}

We conduct two ablation studies on Qwen3.5-9B to assess the contributions of DreamRealm's training data (Table~\ref{tab:realmforge-data-ablation}) and training method (Table~\ref{tab:realmmatch-training-ablation}).

\paragraph{RealmForgePipe data ablation.}
Table~\ref{tab:realmforge-data-ablation} compares four data configurations. RealmSet is the full dataset constructed by RealmForgePipe. Observed Only omits mining for latent perceptual hallucinations and uses only observed hallucinations from greedy decoding to guide image editing. Unguided Edits replaces these hallucination-guided edits with random minimal edits. Removing edited examples altogether yields Original Only, which contains only the original VisCoR-55K data~\citep{scholar_vcstar}. In all three configurations with edits, RealmMatch uses question-centered families containing the original image--answer pair and its edited image--response pairs. For Original Only, we apply SFT to the original image--answer pairs.

\begin{table}[!htb]
\centering
\caption{RealmForgePipe data ablation on Qwen3.5-9B. HallusionBench scores use only image-bearing examples. Bold marks the best value in each column. Subscripts show changes relative to the raw model. Positive values indicate improvement.}
\label{tab:realmforge-data-ablation}\label{tab:ablations}
\begingroup
\fontsize{8.8}{10.4}\selectfont
\setlength{\tabcolsep}{2pt}
\renewcommand{\arraystretch}{1.16}
\resizebox{\textwidth}{!}{%
\begin{tabular}{@{}ll*{8}{c}@{}}
\toprule
 & & \multicolumn{2}{c}{HallusionBench} & \multicolumn{2}{c}{MMVP} & AMBER & \shortstack{MMHal-\\Bench} & \shortstack{CV-Bench-\\2D} & ObjectHal \\
\cmidrule(lr){3-4}\cmidrule(lr){5-6}
\raisebox{0.4ex}[0pt][0pt]{\shortstack[l]{Data\\configuration}} & \raisebox{0.4ex}[0pt][0pt]{\shortstack[l]{Training\\method}} & aAcc $\uparrow$ & fAcc $\uparrow$ & Item Acc $\uparrow$ & Pair Acc $\uparrow$ & Avg Acc $\uparrow$ & Hall.\ Rate $\downarrow$ & Macro Acc $\uparrow$ & CHAIRs $\downarrow$ \\
\midrule
Raw (no training) & ---
 & \resultdelta{71.61}{}
 & \resultdelta{54.91}{}
 & \resultdelta{80.00}{}
 & \resultdelta{63.33}{}
 & \resultdelta{88.34}{}
 & \resultdelta{47.92}{}
 & \resultdelta{82.16}{}
 & \resultdelta{30.33}{} \\
Original Only & SFT
 & \resultdelta{62.15}{\textcolor{red!65!black}{-9.46}}
 & \resultdelta{38.73}{\textcolor{red!65!black}{-16.18}}
 & \resultdelta{80.33}{\textcolor[rgb]{0.05,0.35,0.26}{+0.33}}
 & \resultdelta{62.67}{\textcolor{red!65!black}{-0.66}}
 & \resultdelta{87.21}{\textcolor{red!65!black}{-1.13}}
 & \resultdelta{35.42}{\textcolor[rgb]{0.05,0.35,0.26}{+12.50}}
 & \resultdelta{79.63}{\textcolor{red!65!black}{-2.53}}
 & \resultdelta{11.00}{\textcolor[rgb]{0.05,0.35,0.26}{+19.33}} \\
Unguided Edits & RealmMatch
 & \resultdelta{70.77}{\textcolor{red!65!black}{-0.84}}
 & \resultdelta{51.16}{\textcolor{red!65!black}{-3.75}}
 & \resultdelta{83.67}{\textcolor[rgb]{0.05,0.35,0.26}{+3.67}}
 & \resultdelta{69.33}{\textcolor[rgb]{0.05,0.35,0.26}{+6.00}}
 & \resultdelta{90.97}{\textcolor[rgb]{0.05,0.35,0.26}{+2.63}}
 & \resultdelta{34.38}{\textcolor[rgb]{0.05,0.35,0.26}{+13.54}}
 & \resultdelta{80.37}{\textcolor{red!65!black}{-1.79}}
 & \resultdelta{\textbf{8.00}}{\textcolor[rgb]{0.05,0.35,0.26}{+22.33}} \\
Observed Only & RealmMatch
 & \resultdelta{74.66}{\textcolor[rgb]{0.05,0.35,0.26}{+3.05}}
 & \resultdelta{53.47}{\textcolor{red!65!black}{-1.44}}
 & \resultdelta{86.00}{\textcolor[rgb]{0.05,0.35,0.26}{+6.00}}
 & \resultdelta{74.00}{\textcolor[rgb]{0.05,0.35,0.26}{+10.67}}
 & \resultdelta{90.04}{\textcolor[rgb]{0.05,0.35,0.26}{+1.70}}
 & \resultdelta{30.21}{\textcolor[rgb]{0.05,0.35,0.26}{+17.71}}
 & \resultdelta{81.86}{\textcolor{red!65!black}{-0.30}}
 & \resultdelta{13.33}{\textcolor[rgb]{0.05,0.35,0.26}{+17.00}} \\
\rowcolor{dreamrealmshade}
\textbf{RealmSet} & RealmMatch
 & \resultdelta{\textbf{80.02}}{\textcolor[rgb]{0.05,0.35,0.26}{+8.41}}
 & \resultdelta{\textbf{60.98}}{\textcolor[rgb]{0.05,0.35,0.26}{+6.07}}
 & \resultdelta{\textbf{86.67}}{\textcolor[rgb]{0.05,0.35,0.26}{+6.67}}
 & \resultdelta{\textbf{74.67}}{\textcolor[rgb]{0.05,0.35,0.26}{+11.34}}
 & \resultdelta{\textbf{91.01}}{\textcolor[rgb]{0.05,0.35,0.26}{+2.67}}
 & \resultdelta{\textbf{27.08}}{\textcolor[rgb]{0.05,0.35,0.26}{+20.84}}
 & \resultdelta{\textbf{82.89}}{\textcolor[rgb]{0.05,0.35,0.26}{+0.73}}
 & \resultdelta{8.67}{\textcolor[rgb]{0.05,0.35,0.26}{+21.66}} \\
\bottomrule
\end{tabular}%
}
\endgroup
\end{table}

Table~\ref{tab:realmforge-data-ablation} shows that removing the edited examples derived from latent perceptual hallucination mining yields Observed Only, which performs worse than RealmSet on all eight metrics but still substantially outperforms Raw on six. This suggests that edits guided by observed hallucinations benefit the model and that mining latent perceptual hallucinations provides additional value. With latent hallucination mining, the fraction of HallusionBench images for which Qwen3.5-9B answers every associated question correctly rises from 53.47\% to 60.98\% (fAcc, +7.51 percentage points). Replacing edits guided by observed hallucinations with random edits yields Unguided Edits, which performs worse than Observed Only on six metrics. Unguided Edits no longer outperforms Raw on HallusionBench, but it still improves on Raw on five metrics and achieves the lowest ObjectHal CHAIRs (8.00\%). These results suggest that random edits can improve performance, although their overall benefit is smaller than that of edits guided by the target model's own hallucinations. Finally, Original Only, which applies SFT to the original VisCoR-55K data, does not yield consistent improvements across benchmarks for Qwen3.5-9B.

\paragraph{RealmMatch training-method ablation.}
Table~\ref{tab:realmmatch-training-ablation} compares our proposed RealmMatch method with three widely used post-training methods (SFT, DPO, and GRPO), all trained on the same RealmSet data.

\begin{table}[!htb]
\centering
\caption{RealmMatch training-method ablation on Qwen3.5-9B. All trained variants use the same initialization and RealmSet, including both original and edited examples. HallusionBench scores use only image-bearing examples. Bold marks the best value in each column. Subscripts show changes relative to the raw model. Positive values indicate improvement.}
\label{tab:realmmatch-training-ablation}
\begingroup
\fontsize{8.8}{10.4}\selectfont
\setlength{\tabcolsep}{2pt}
\renewcommand{\arraystretch}{1.16}
\resizebox{\textwidth}{!}{%
\begin{tabular}{@{}ll*{8}{c}@{}}
\toprule
 & & \multicolumn{2}{c}{HallusionBench} & \multicolumn{2}{c}{MMVP} & AMBER & \shortstack{MMHal-\\Bench} & \shortstack{CV-Bench-\\2D} & ObjectHal \\
\cmidrule(lr){3-4}\cmidrule(lr){5-6}
\raisebox{0.4ex}[0pt][0pt]{\shortstack[l]{Training\\method}} & \raisebox{0.4ex}[0pt][0pt]{\shortstack[l]{Data\\configuration}} & aAcc $\uparrow$ & fAcc $\uparrow$ & Item Acc $\uparrow$ & Pair Acc $\uparrow$ & Avg Acc $\uparrow$ & Hall.\ Rate $\downarrow$ & Macro Acc $\uparrow$ & CHAIRs $\downarrow$ \\
\midrule
Raw (no training) & ---
 & \resultdelta{71.61}{}
 & \resultdelta{54.91}{}
 & \resultdelta{80.00}{}
 & \resultdelta{63.33}{}
 & \resultdelta{88.34}{}
 & \resultdelta{47.92}{}
 & \resultdelta{82.16}{}
 & \resultdelta{30.33}{} \\
SFT & RealmSet
 & \resultdelta{74.66}{\textcolor[rgb]{0.05,0.35,0.26}{+3.05}}
 & \resultdelta{54.34}{\textcolor{red!65!black}{-0.57}}
 & \resultdelta{83.67}{\textcolor[rgb]{0.05,0.35,0.26}{+3.67}}
 & \resultdelta{68.67}{\textcolor[rgb]{0.05,0.35,0.26}{+5.34}}
 & \resultdelta{\textbf{92.12}}{\textcolor[rgb]{0.05,0.35,0.26}{+3.78}}
 & \resultdelta{35.42}{\textcolor[rgb]{0.05,0.35,0.26}{+12.50}}
 & \resultdelta{82.85}{\textcolor[rgb]{0.05,0.35,0.26}{+0.69}}
 & \resultdelta{12.33}{\textcolor[rgb]{0.05,0.35,0.26}{+18.00}} \\
DPO & RealmSet
 & \resultdelta{69.09}{\textcolor{red!65!black}{-2.52}}
 & \resultdelta{50.00}{\textcolor{red!65!black}{-4.91}}
 & \resultdelta{86.00}{\textcolor[rgb]{0.05,0.35,0.26}{+6.00}}
 & \resultdelta{74.00}{\textcolor[rgb]{0.05,0.35,0.26}{+10.67}}
 & \resultdelta{89.46}{\textcolor[rgb]{0.05,0.35,0.26}{+1.12}}
 & \resultdelta{34.38}{\textcolor[rgb]{0.05,0.35,0.26}{+13.54}}
 & \resultdelta{78.81}{\textcolor{red!65!black}{-3.35}}
 & \resultdelta{11.33}{\textcolor[rgb]{0.05,0.35,0.26}{+19.00}} \\
GRPO & RealmSet
 & \resultdelta{75.39}{\textcolor[rgb]{0.05,0.35,0.26}{+3.78}}
 & \resultdelta{56.36}{\textcolor[rgb]{0.05,0.35,0.26}{+1.45}}
 & \resultdelta{84.33}{\textcolor[rgb]{0.05,0.35,0.26}{+4.33}}
 & \resultdelta{70.00}{\textcolor[rgb]{0.05,0.35,0.26}{+6.67}}
 & \resultdelta{90.65}{\textcolor[rgb]{0.05,0.35,0.26}{+2.31}}
 & \resultdelta{30.21}{\textcolor[rgb]{0.05,0.35,0.26}{+17.71}}
 & \resultdelta{82.16}{0.00}
 & \resultdelta{15.67}{\textcolor[rgb]{0.05,0.35,0.26}{+14.66}} \\
\rowcolor{dreamrealmshade}
\textbf{RealmMatch} & RealmSet
 & \resultdelta{\textbf{80.02}}{\textcolor[rgb]{0.05,0.35,0.26}{+8.41}}
 & \resultdelta{\textbf{60.98}}{\textcolor[rgb]{0.05,0.35,0.26}{+6.07}}
 & \resultdelta{\textbf{86.67}}{\textcolor[rgb]{0.05,0.35,0.26}{+6.67}}
 & \resultdelta{\textbf{74.67}}{\textcolor[rgb]{0.05,0.35,0.26}{+11.34}}
 & \resultdelta{91.01}{\textcolor[rgb]{0.05,0.35,0.26}{+2.67}}
 & \resultdelta{\textbf{27.08}}{\textcolor[rgb]{0.05,0.35,0.26}{+20.84}}
 & \resultdelta{\textbf{82.89}}{\textcolor[rgb]{0.05,0.35,0.26}{+0.73}}
 & \resultdelta{\textbf{8.67}}{\textcolor[rgb]{0.05,0.35,0.26}{+21.66}} \\
\bottomrule
\end{tabular}%
}
\endgroup
\end{table}

On RealmSet, SFT, DPO, and GRPO each outperform Raw on most metrics, indicating that RealmSet also provides effective supervision for these methods. On HallusionBench, RealmMatch achieves an aAcc of 80.02\%, outperforming the next-best method, GRPO (75.39\%), by 4.63 percentage points. These results support the effectiveness of RealmMatch for hallucination assignment. Within each RealmSet family, the question stays fixed, and each of the target model's hallucinated responses is paired with a distinct edited image that makes it correct. Bidirectional contrast trains the model to suppress each such response on the original image and assign it to its supporting image. With this joint supervision, RealmMatch outperforms SFT, DPO, and GRPO on seven of eight metrics. Appendix~\ref{app:loss-ablation} outlines further ablations of the individual RealmMatch loss terms.

\endgroup

\FloatBarrier
\section{Conclusion}
\label{sec:conclusion}
We introduce DreamRealm, a post-training pipeline based on hallucination assignment. DreamRealm mines a target model's own hallucinations and teaches the model to associate each hallucinated response with the visual conditions that make it correct. Using these hallucinations to guide image editing, we constructed RealmSet, which contains 43,483 human-approved edited image--response pairs spanning 22 types of visual perception tasks. Experiments on three target models across six visual perception and hallucination benchmarks show that DreamRealm improves visual perception and mitigates hallucinations. With Qwen3.5-9B as its backbone, DreamRealm outperforms existing hallucination mitigation methods on seven of eight metrics. Our ablation studies highlight the importance of mining latent perceptual hallucinations and show that RealmMatch makes more effective use of the image--response correspondences in RealmSet. Appendix~\ref{sec:limitations} discusses the limitations. Together, these findings suggest that \textbf{hallucination mining and assignment offer a promising post-training paradigm for improving visual perception in vision--language models.}

\clearpage
{\small
\bibliographystyle{ieeenat_fullname}
\bibliography{main}
}

\clearpage
\pdfbookmark[1]{Supplementary Material}{supplementary}
{\centering\sffamily\bfseries\fontsize{16}{20}\selectfont\color{reportaccent}Supplementary Material\par}
\vspace{12pt}
\appendix
\begingroup
\setlength{\parskip}{3pt plus 1pt}
\titlespacing*{\paragraph}{0pt}{6pt}{0.8em}
\section*{Contents of the Supplementary Materials}

Appendix~A describes the data pipeline for hallucination materialization. Appendices~B and C show examples of hallucination mining and hallucination-guided image editing, respectively. Appendix~D describes RealmSet construction, the distribution of question types, and the human review protocol. Appendix~E walks through a complete hallucination-guided image editing example. Appendices~F and G describe the training settings and evaluation setup and scoring protocols, respectively. Appendix~H presents qualitative comparisons with the base model. Appendix~I documents the base models and evaluation setups for the compared methods. Appendix~J reports the RealmMatch loss ablations. Appendix~K discloses LLM use. Appendix~\ref{sec:limitations} discusses limitations.

\section{Hallucination Materialization Algorithm}
\label{app:materialization}

Algorithm~\ref{alg:materialization} shows the per-edit synthesis and verification procedure described in Section~\ref{sec:hallucination-materialization}, including difficulty tagging, quality verification, question-dependent blind checking, and bounded retry logic. Stages 1.5 and 2.5 record metadata only and never reject samples. Each instruction--type pair permits at most two synthesis attempts in total. Human annotators review individual edits in the first round and every family in the second round before post-processing.

\paragraph{Minimal-editing principle.}
Minimal editing prioritizes changes necessary to make the target response valid while preserving unrelated scene content and structure. The planner specifies preservation and modification regions $R_{pres}$ and $R_{chg}$ and orders three instructions by increasing expected edit extent. The verifier evaluates minimal editing on a five-point scale, with $s_{min}\geq4$ required for acceptance, together with $s_{es}\geq4$ and $s_{pl}\geq3$. This is a planning preference and a verifier-based quality constraint, rather than minimization of a pixel-distance objective or a guarantee of a globally minimal edit. Each instruction can fall back to full-image regeneration, as shown in Algorithm~\ref{alg:materialization}.

\paragraph{Triage subroutine (Stage 1).}
The verifier first answers the original question without access to either supplied answer; counting questions require counting objects individually. It then determines question validity ($A$), agreement with the ground-truth answer ($B$), equivalence between the target response and the ground-truth answer ($C$), refusal ($D$), and the error type $e$. In optional two-pass auditing, both passes must satisfy $A$ and $B$, while either pass triggering $C$ or $D$ causes rejection; the stricter audit judgment is retained.

\paragraph{Blind-check subroutine (Stage 4b).}
The algorithm explicitly distinguishes three modes. Strict mode obtains an answer from the edited image with the target response hidden, then judges factual equivalence to $y_m$. Relaxed mode extracts a target visual feature from $(R_{chg},y_m)$ and accepts only a YES response to its presence in the edited image. Skip mode performs no additional check and returns true. Only strict-mode check failures retry the same instruction--type pair; failed synthesis also consumes an attempt and retries, whereas a quality-threshold failure or a relaxed-mode failure advances to the next editing type. All verification scores use a five-point scale.

\paragraph{Within-family deduplication and verification.}
In the second round of human review, annotators examine every family, including the original pair and the edits for the same question. They check whether each image supports its assigned response and whether any other response in the family also applies to that image. They remove affected pairs when they find a cross-match, applying the acceptance criterion in Eq.~\ref{eq:family-exclusivity}. After human review, we remove duplicate images and keep one representative pair per distinct response meaning, giving priority to the original pair.

\par\endgroup
\clearpage
\begin{algorithm}[H]
\caption{Verified hallucination materialization in RealmForgePipe}
\label{alg:materialization}
\small
\begin{algorithmic}[1]
\Require $(I,q,y^\star,y_m)$; verifier $\mathcal{V}$; editor $\mathcal{E}$ (\texttt{gpt-image-2})
\Require $(\tau_{es},\tau_{pl},\tau_{min})=(4,3,4)$; $R_{ins}=3$; $R_{syn}=2$
\Ensure Candidate record $(I^\star,q,y^\star\!\to\!y_m,\text{metadata})$, or \textsc{drop}/\textsc{fail}
\State \textbf{if} $I$ unavailable \textbf{then return} \textsc{drop}
\State $\hat y\gets\mathcal{V}(I,q)$ \Comment{1. Triage: neither supplied answer is shown}
\State $(A,B,C,D,e)\gets\Call{Audit}{q,\hat y,y^\star,y_m}$
\Statex \hspace{\algorithmicindent} $A$: well-defined question; $B$: $\hat y\equiv y^\star$; $C$: $y_m\equiv y^\star$; $D$: refusal; $e$: error type
\State \textbf{if} $\neg A\lor\neg B\lor C\lor D$ \textbf{then return} \textsc{drop}
\State $(d,h)\gets\Call{AssessDifficulty}{\mathcal{V},I,q,y_m,e}$ \Comment{1.5. Record only; never reject}
\Statex \hspace{\algorithmicindent} $d\in\{1,\ldots,5\}$: editing/verification difficulty (higher is harder); $h$: optional difficulty tags
\State $(t_0,R_{pres},R_{chg},\langle i_1,i_2,i_3\rangle,p_{t2i})\gets\Call{Plan}{\mathcal{V},I,q,y^\star,y_m,e}$ \Comment{2. Plan}
\State $hard\_case\gets[t_0=\textsc{T2IReplan}\land e=\textsc{PerceptionLimit}]$ \Comment{2.5. Record only}
\Statex \hspace{\algorithmicindent} $t_0$: local editing, global editing, or text-to-image replanning; $i_1,i_2,i_3$: increasing edit extent
\State $\mathcal{T}\gets\mathrm{Dedup}(\langle t_0,\textsc{T2IReplan}\rangle)$ \Comment{Preserve fallback order}
\For{$k=1,\ldots,R_{ins}$}
  \For{$t\in\mathcal{T}$}
    \For{$a=1,\ldots,R_{syn}$}
      \State $I'\gets\Call{Synth}{\mathcal{E},I,\mathrm{Wrap}(i_k,e),t,p_{t2i}}$ \Comment{3. Synthesis}
      \State \textbf{if} $I'=\emptyset$ \textbf{then continue} \Comment{Next synthesis attempt}
      \State $(s_{es},s_{pl},s_{min})\gets\Call{Verify}{\mathcal{V},I,I',q,y_m,R_{pres}}$ \Comment{4. Quality verification}
      \State \textbf{if} $s_{es}<\tau_{es}\lor s_{pl}<\tau_{pl}\lor s_{min}<\tau_{min}$ \textbf{then break}
      \State $mode\gets\Call{Classify}{q,y_m}\in\{\text{strict},\text{relaxed},\text{skip}\}$ \Comment{4b. Blind-check gate}
      \If{$mode=\text{strict}$}
        \State $\tilde y\gets\mathcal{V}(I',q)$ \Comment{Target answer is hidden during answering}
        \State $m\gets\Call{Equiv}{\mathcal{V},\tilde y,y_m}$
      \ElsIf{$mode=\text{relaxed}$}
        \State $f\gets\Call{ExtractFeature}{\mathcal{V},R_{chg},y_m}$
        \State $m\gets[\mathcal{V}(I',\text{``Is feature }f\text{ visibly present?''})=\text{YES}]$
      \Else
        \State $m\gets\mathrm{true}$ \Comment{Skip the additional check}
      \EndIf
      \If{$m$}
        \State $meta\gets\{i_k,t,s_{es},s_{pl},s_{min},mode,d,h,hard\_case\}$
        \State \Return $(I',q,y^\star\!\to\!y_m,meta)$ \Comment{Candidate for human review}
      \EndIf
      \State \textbf{if} $mode\neq\text{strict}$ \textbf{then break} \Comment{Only strict failures retry this pair}
    \EndFor
  \EndFor
\EndFor
\State \Return \textsc{fail}
\Statex \textbf{After per-edit verification:} Review each candidate image--response pair for correctness.
\Statex Review every family with its original pair for correctness and mutual exclusivity.
\Statex Remove cross-matches under Eq.~\ref{eq:family-exclusivity}, then deduplicate images and responses.
\end{algorithmic}
\end{algorithm}

\clearpage
\section{Qualitative Examples of Hallucination Mining}
\label{app:mining-examples}

\begin{figure}[H]
  \centering
  \includegraphics[width=\linewidth,height=0.85\textheight,keepaspectratio]{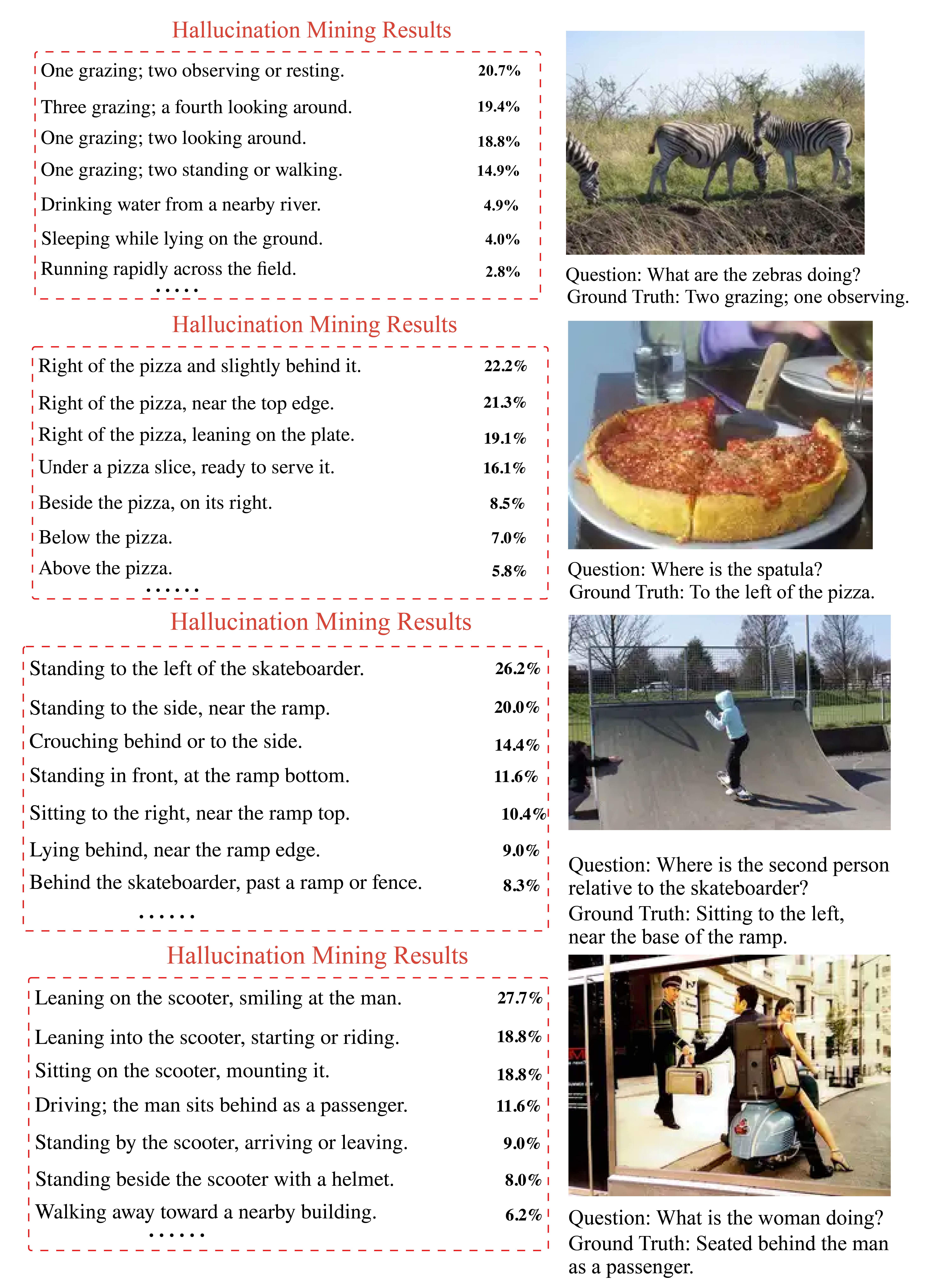}
  \caption{Hallucination mining results from RealmForgePipe on four examples involving actions and spatial relations. Each example shows the original image, question, ground-truth answer, and selected mined responses ranked by likelihood. Percentages denote candidate-relative weight $\rho_i(e)$ within the full error candidate pool, rather than calibrated answer probabilities; displayed values are not renormalized over the shown subset. Responses are shortened for readability.}
  \label{fig:hallucination-mining-examples}
\end{figure}

\clearpage
\begin{figure}[H]
  \centering
  \includegraphics[width=\linewidth,height=0.92\textheight,keepaspectratio]{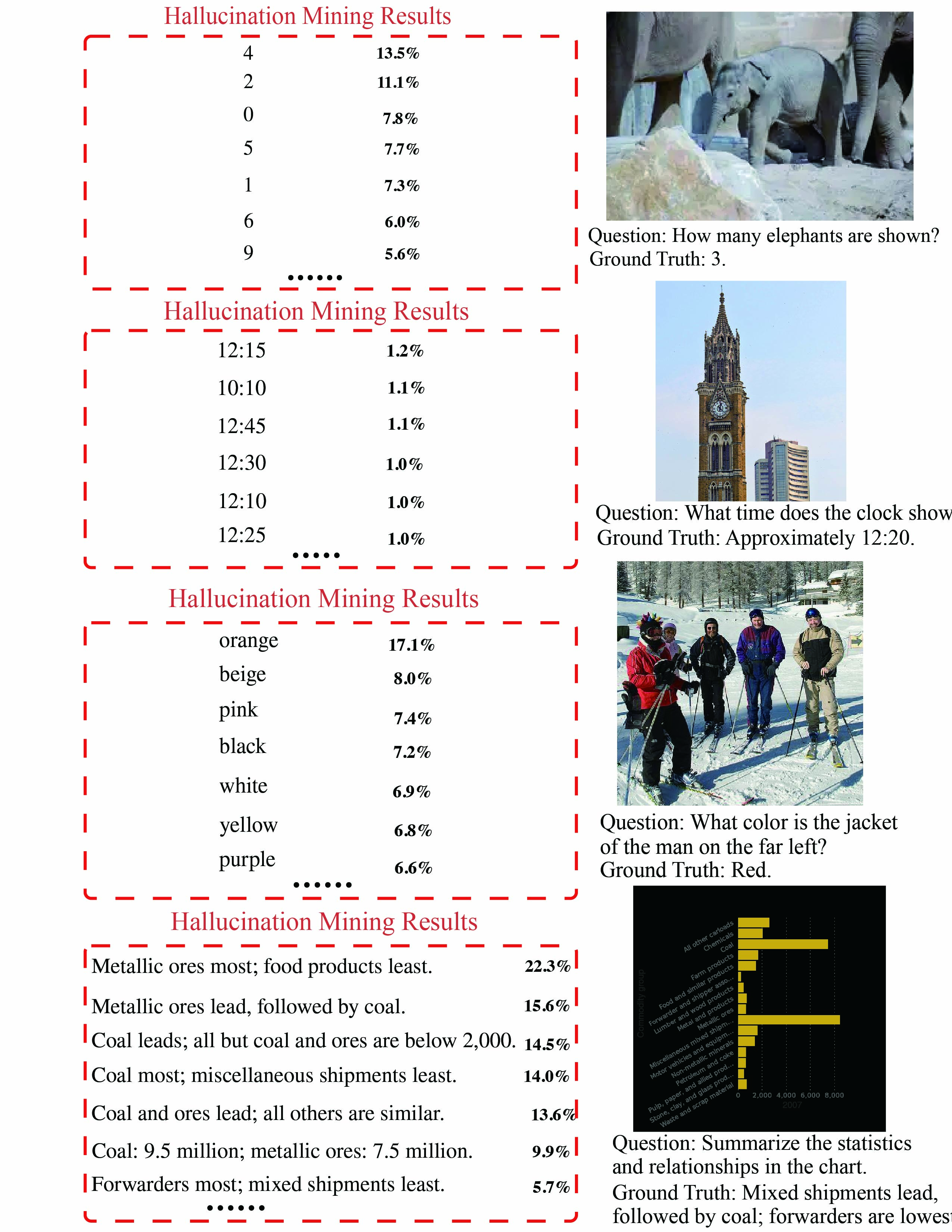}
  \caption{Additional hallucination mining examples covering counting, clock reading, color recognition, and chart analysis. Selected candidates are shown with their candidate-relative weights from the original mining runs, without renormalization over the displayed subset. These weights are not calibrated answer probabilities. Chart responses are shortened for readability.}
  \label{fig:hallucination-mining-additional}
\end{figure}

\clearpage
\section{Hallucination-Guided Image Editing Examples}
\label{app:editing-examples}

\begin{figure}[H]
  \centering
  \includegraphics[width=\linewidth,height=0.86\textheight,keepaspectratio]{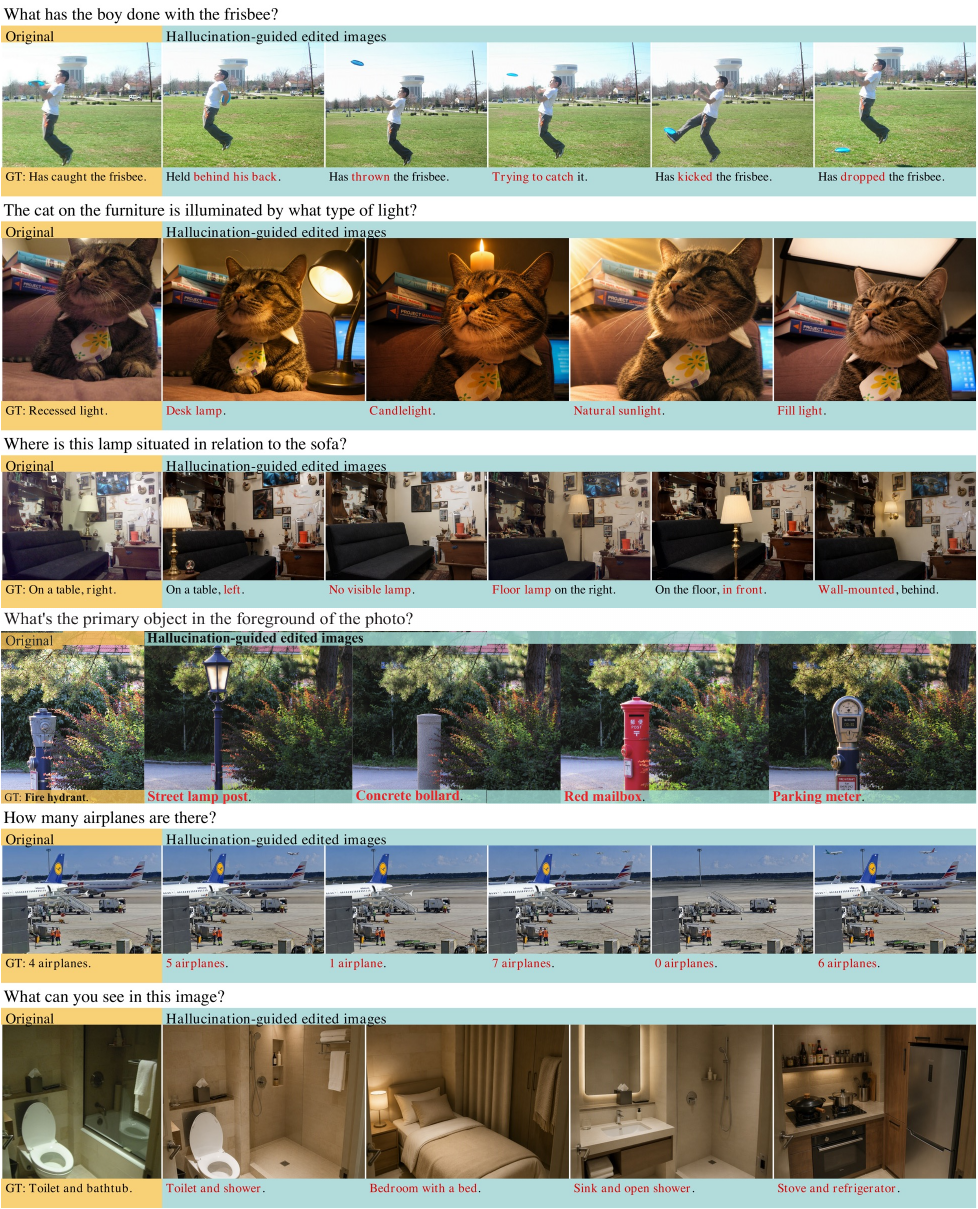}
  \caption{Hallucination-guided image editing examples from RealmForgePipe (1/2). From top to bottom: actions, illumination, spatial relations, foreground object identification, object counts, and scene structure. Each row pairs the original image and ground-truth answer (yellow) with edited images and their corresponding hallucinated responses (cyan). Red text highlights the response attributes targeted by editing. Responses are shortened for readability.}
  \label{fig:hallucination-editing-examples}
\end{figure}

\clearpage
\begin{figure}[H]
  \centering
  \includegraphics[width=\linewidth,height=0.87\textheight,keepaspectratio]{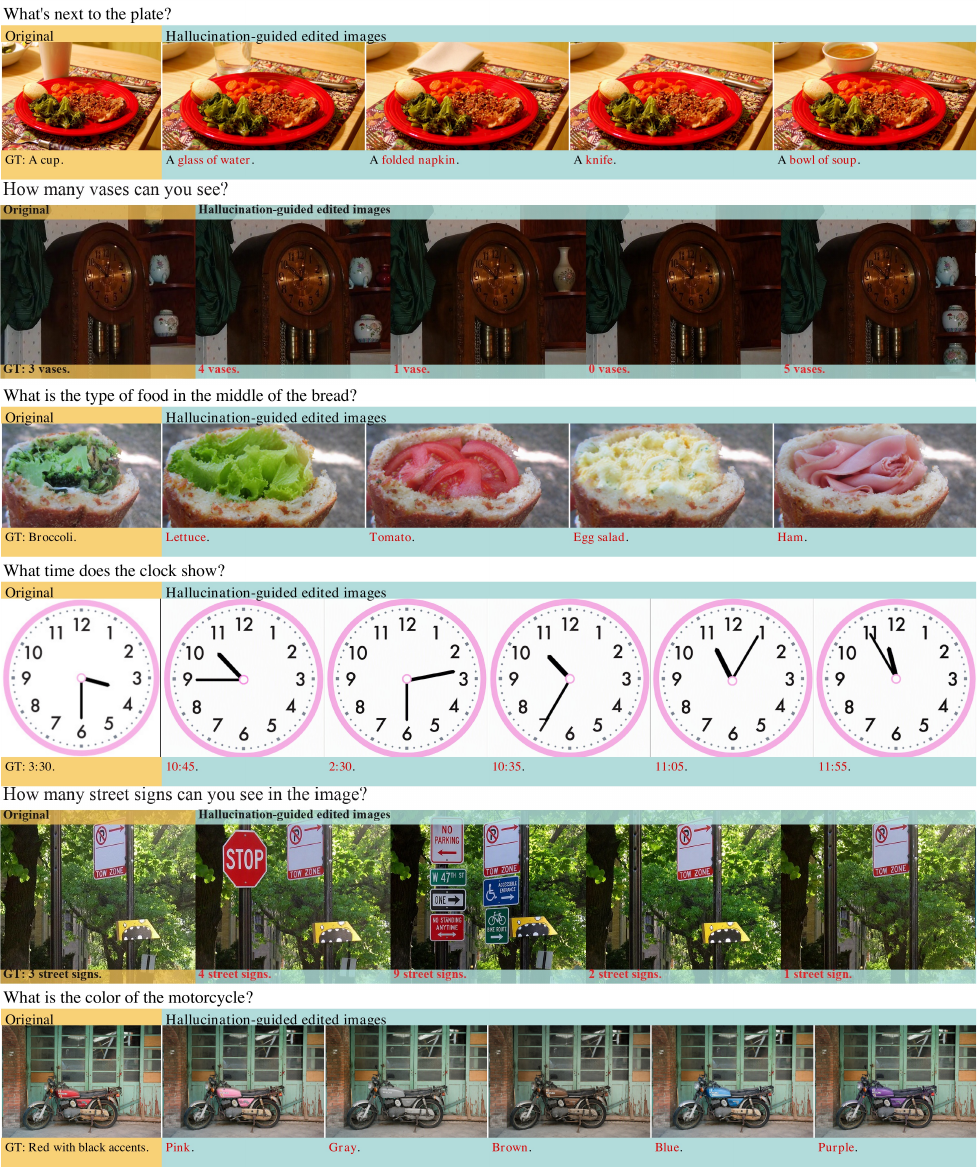}
  \caption{Hallucination-guided image editing examples from RealmForgePipe (2/2). From top to bottom: local object replacement, vase counts, food content, clock time, street-sign counts, and color attributes. Together with Figure~\ref{fig:hallucination-editing-examples}, the two pages show 12 original-image families and 53 edited images. Yellow and cyan labels denote original and edited image--response pairs; red text highlights the targeted attributes.}
  \label{fig:hallucination-editing-spatial}
\end{figure}

\clearpage

\begingroup
\setlength{\parskip}{2.6pt plus 1pt}
\titlespacing*{\paragraph}{0pt}{5pt}{0.65em}
\section{RealmSet Construction and Question Types}
\label{app:data-construction}

\paragraph{Construction across three batches.}
Table~\ref{tab:data-construction} follows candidate edit variants from hallucination mining to human review across the three construction batches used for the totals in Section~\ref{sec:realmset}. The historical batch draws on the VisCoR-55K VQA data released with VC-STaR~\citep{scholar_vcstar}. A question can contribute several candidates, and each approved image--response pair counts once. The historical mining and editing totals include records reconstructed from batch summaries.

\begin{table}[H]
\centering
\small
\caption{RealmSet construction counts by batch. Counts are cumulative within each batch.}
\label{tab:data-construction}
\begin{tabular}{@{}lrrrr@{}}
\toprule
Stage & Historical & Batch 1 & Batch 2 & Total \\
\midrule
Retained and scored candidates & 487,812 & 125,493 & 19,946 & 633,251 \\
Selected candidates & 114,982 & 36,662 & 18,300 & 169,944 \\
Candidates entering synthesis & 96,320 & 31,048 & 15,279 & 142,647 \\
Successful edits & 91,986 & 30,131 & 14,980 & 137,097 \\
Human-approved edited pairs & 28,276 & 7,320 & 7,887 & 43,483 \\
\bottomrule
\end{tabular}
\end{table}

\paragraph{Editing yield and record coverage.}
Of the 142,647 candidates that entered synthesis, 43,483 yielded a human-approved edited image--response pair (30.48\%). Of the 137,097 successful edits, 100,767 have distinct recorded image paths. Historical batch summaries account for the remaining 36,330. All 43,483 approved pairs have distinct source IDs and edited image paths. The consolidated blind-check labels were either imputed or inferred from human approval, so the records do not establish an empirical blind-check pass rate.

\paragraph{Qwen3.5-9B training data.}
The Qwen3.5-9B training data include 27,038 edited examples across 12,340 original-image families. Appendix~\ref{app:training-settings} describes the training setup.

\subsection{Distribution by Question Type}
\label{app:data-categories}

We assigned each question a single type based on the task it asks the model to perform. We first applied rules to the question text, then used Qwen3.7-Max to classify unresolved questions and check a sample from each rule-assigned category. We then reviewed all distinct questions labeled as counting. The geometric figures and structures category includes 2,238 pairs involving counts of figures and 151 involving geometric structure, such as the number of faces or vertices. Questions that restrict the count by an attribute, such as the color of a triangle, fall under counting by attribute. Chart interpretation covers questions about specific chart values and questions about overall patterns or trends.

\begin{table}[H]
\centering
\small
\caption{Human-approved edited image--response pairs in RealmSet by question type. Each pair is counted once. Percentages are relative to all 43,483 approved pairs.}
\label{tab:data-categories}
\begin{tabular}{@{}lrr@{\hspace{2em}}lrr@{}}
\toprule
Question type & Pairs & Share (\%) & Question type & Pairs & Share (\%) \\
\midrule
General counting & 13,235 & 30.44 & Shape & 1,028 & 2.36 \\
Time reading & 4,643 & 10.68 & Spatial position & 980 & 2.25 \\
Counting by attribute & 4,313 & 9.92 & Measurement & 896 & 2.06 \\
Chart interpretation & 3,863 & 8.88 & Action & 756 & 1.74 \\
Geometric figures and structures & 2,389 & 5.49 & Material & 516 & 1.19 \\
Text recognition & 2,092 & 4.81 & Size & 387 & 0.89 \\
Color & 1,748 & 4.02 & Comparison & 295 & 0.68 \\
Reasoning & 1,603 & 3.69 & Multi-image reasoning & 252 & 0.58 \\
Existence & 1,471 & 3.38 & Sentiment & 236 & 0.54 \\
Identification & 1,300 & 2.99 & Probability & 138 & 0.32 \\
Scene description & 1,293 & 2.97 & Other & 49 & 0.11 \\
\midrule
\multicolumn{4}{@{}l}{\textbf{Total}} & \textbf{43,483} & \textbf{100.00} \\
\bottomrule
\end{tabular}
\end{table}

\paragraph{Image editing configuration and cost.}
We use GPT-Image-2 with the quality parameter set to \texttt{low}. Image generation costs CNY~0.35 per generated image, excluding VLM verification and human annotation. This is a per-generation cost, not a cost per accepted edited pair: retries and generated images that fail subsequent checks also contribute to construction cost. The human annotation protocol and service price are described in Appendix~\ref{app:human-review}.

\subsection{Human Annotation Review Protocol}
\label{app:human-review}

\paragraph{Annotators and task.}
Human review is conducted by 12 annotators, all from professional data annotation companies. Before annotation begins, the task and decision criteria are communicated to the annotators. Given an edited image and its proposed response, each annotator judges whether the response is supported by the image. Samples for which this judgment is uncertain are discarded. Each sample is initially reviewed by one annotator, and annotators work on separate, non-overlapping batches.

\paragraph{Quality control.}
In the second round, reviewers check every family. For each family, reviewers examine the original and edited images together with their assigned responses to the same question. They check that each image supports its assigned response and that the response does not also apply to another image in the family. If reviewers find such a cross-match, they remove the affected pair. Semantically equivalent candidate responses have already been removed during hallucination mining, and reviewers rarely encountered cross-matches in practice. If an annotator fails this quality check, the data assigned to that annotator are returned and reassigned to a different annotator for reannotation. After human review, we deduplicate the retained images and responses within each family.

\paragraph{Throughput and cost.}
The review throughput is 500 samples per annotator-hour, and the annotation service is charged at CNY~0.0568 per sample.

\section{A Worked Example of Hallucination-Guided Editing}
\label{app:editing-walkthrough}

We trace the recorded automated editing run for case \texttt{orig\_148}, variant 0. The question is \emph{What is the size of the red block?} The original answer is \emph{Small}; the target hallucinated response is \emph{Large}. This trace contains one target response and three alternative editing instructions for that response.

\begin{figure}[H]
\centering
\begin{minipage}{0.39\linewidth}
\centering
\includegraphics[width=\linewidth]{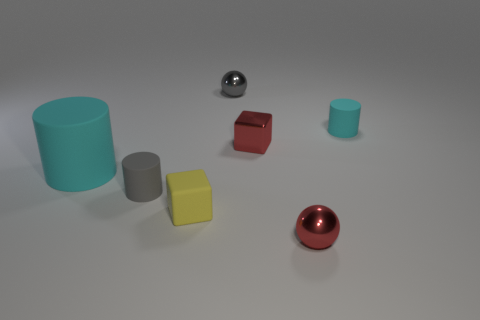}\\
\small Original image: \textbf{Small}
\end{minipage}\hspace{.035\linewidth}
\begin{minipage}{0.39\linewidth}
\centering
\includegraphics[width=\linewidth]{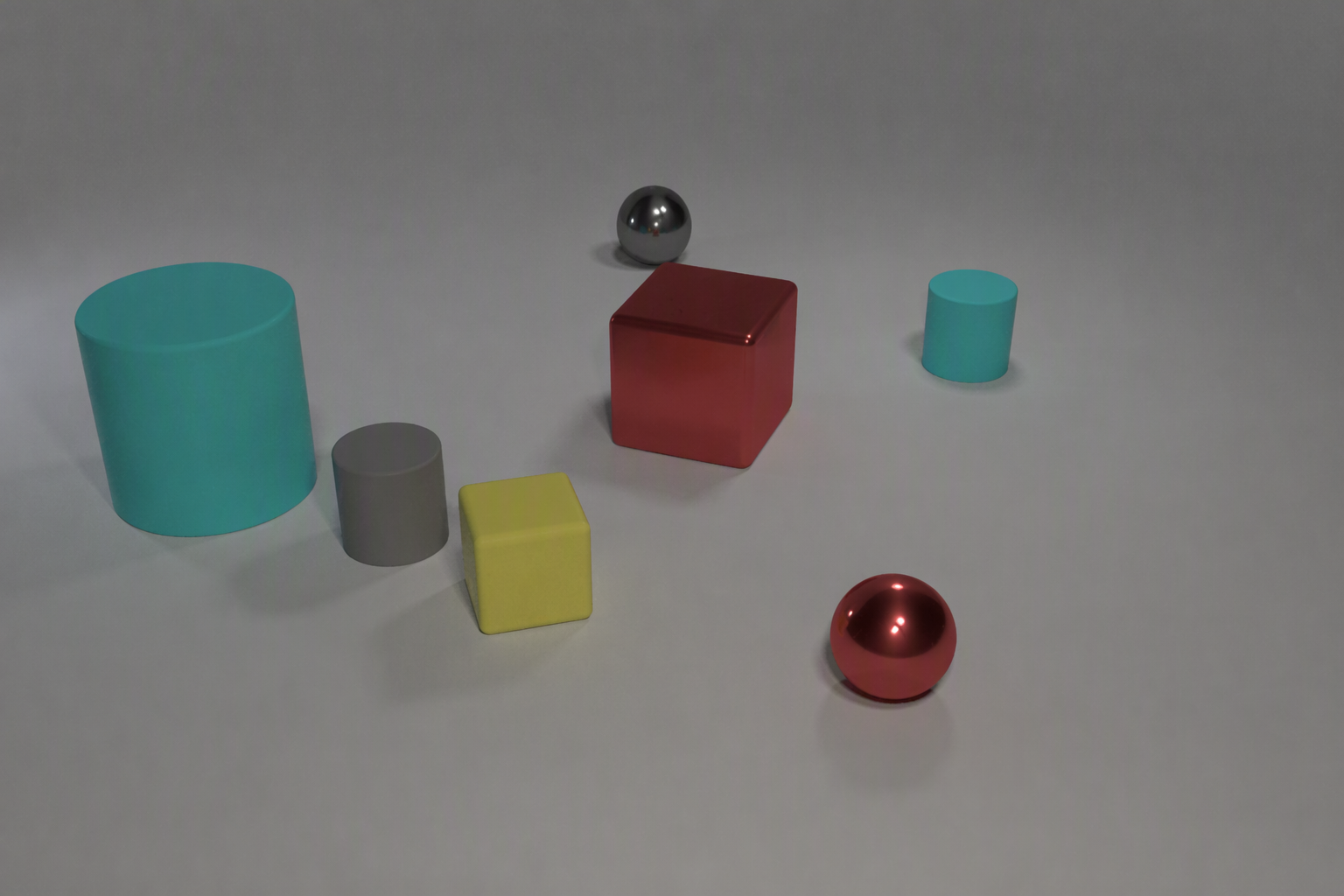}\\
\small Edited image: \textbf{Large}
\end{minipage}
\caption{A recorded editing example. Enlarging the red cube makes the target response \emph{Large} valid for the same question.}
\label{fig:editing-walkthrough}
\end{figure}

\paragraph{Screening and difficulty assessment.}
The verifier (\texttt{gemini-2.5-pro-06-17}; Gemini 2.5~\citep{scholar_gemini25}) returns \emph{Small}, marks the question and ground-truth answer as valid, and identifies an attribute error. It assigns difficulty 4/5 with shape-preservation, shadow-manipulation, and reflection-manipulation tags. These tags do not block editing. 

\paragraph{Planning.}
The verifier selects local editing, identifies the red cube as the object to modify, and specifies that the background, lighting, and other objects should be preserved. It proposes three instructions:
\begin{enumerate}
\setlength{\itemsep}{0pt}
\setlength{\parskip}{0pt}
\item Increase the size of the red cube.
\item Scale up the red cube to be approximately twice its current size.
\item Make the red cube a \emph{large} object by increasing its size so its height is comparable to the large cyan cylinder on the left, and adjust its shadow accordingly.
\end{enumerate}

\paragraph{Execution and automated verification.}
GPT-Image-2 (\texttt{gpt-image-2-0421-global}) executes the first instruction, prefixed with \texttt{[Error type: attribute]}. Generation takes 22.746 seconds. The verifier assigns 5/5 to each of edit success, visual plausibility, and minimal editing, exceeding the respective thresholds of 4, 3, and 4. These are the verifier's reported scores.

\paragraph{Blind check and recorded outcome.}
Given only the edited image and the question, the verifier answers that the red block is \emph{large}. A subsequent equivalence check against \emph{Large} returns \texttt{YES}. The run stores the edited image and metadata with status \texttt{ok}. The first generation under the first instruction succeeds, so neither alternative instruction nor the full-redraw fallback is executed. The supplied trace ends at automated storage and contains no human-review decision; final dataset acceptance remains subject to the human-review protocol in Appendix~\ref{app:human-review}.

\section{Training Settings}
\label{app:training-settings}

\paragraph{Training data and model configurations.}
For Qwen3.5-9B, we compare four training data configurations (Table~\ref{tab:realmforge-data-ablation}) and four training methods (Table~\ref{tab:realmmatch-training-ablation}). We use Qwen3.5-9B as both the hallucination-mining model and the primary training model. We train on original examples and the 27,038 HG-Edited training examples in RealmSet (Section~\ref{sec:realmset}).
For Qwen2.5-VL-7B~\citep{scholar_qwen25vl} and InternVL3.5-8B~\citep{scholar_internvl35}, we separately mine hallucinations from each target model and construct its corresponding training data with RealmForgePipe. Each model is then trained with RealmMatch on its own hallucination-derived data; the Qwen3.5-9B dataset counts above do not describe these separate datasets.

\paragraph{Model adaptation.}
For Qwen3.5-9B, we initialize from the original pretrained checkpoint and optimize the RealmMatch objective in Section~\ref{sec:realmmatch}. We update the language model and multimodal projector while freezing the vision tower, without using LoRA adapters. This configuration has 8,996,576,512 trainable parameters. We conducted the experiments on 16 NVIDIA H100 GPUs. The saved training configuration for the Qwen3.5-9B checkpoint evaluated in Table~1 records eight GPU workers. The run used BF16 precision, gradient checkpointing, and DeepSpeed ZeRO-3 with CPU optimizer offloading. Dropout is disabled. The training hyperparameters are summarized in Table~\ref{tab:training-settings}.

\begin{table}[H]
\centering
\small
\caption{Training hyperparameters for the Qwen3.5-9B RealmMatch configuration. Batch units are complete question-centered families, rather than individual image--response combinations.}
\label{tab:training-settings}
\setlength{\tabcolsep}{3pt}
\begin{tabular*}{\linewidth}{@{\extracolsep{\fill}}lrlr@{}}
\toprule
Hyperparameter & Value & Hyperparameter & Value \\
\cmidrule(r){1-2}\cmidrule(l){3-4}
Hardware reported in Section~4.1 & 16 NVIDIA H100 & Maximum sequence length & 6,144 tokens \\
Training epochs & 3 & Image processor pixel budget & 262,144 \\
Peak learning rate & $5\times10^{-6}$ & Contrastive temperature $\tau$ & 1.0 \\
Learning-rate schedule & Cosine & Response contrast weight $\lambda_r$ & 0.5 \\
Warmup ratio & 0.03 & Image contrast weight $\lambda_c$ & 0.5 \\
Weight decay & 0 & SFT loss coefficient & 1.0 \\
Maximum gradient norm & 1.0 & Precision & BF16 \\
Per-device micro-batch & 1 family & Checkpoint evaluated in Table~1 & Epoch 2 (step 648) \\
Gradient accumulation & 10 micro-steps & Checkpoint saving & Every epoch \\
Effective batch size & 80 families &  &  \\
\bottomrule
\end{tabular*}
\end{table}

\paragraph{Family-wise computation.}
Each training micro-step contains one complete family and uses the loss defined
in Section~\ref{sec:realmmatch}: edited families use SFT and bidirectional
contrast, whereas families without accepted edits use SFT and response
contrast. Matrix cells are processed in micro-batches of at most four cells
and 24,576 padded input tokens. Families that fit in one micro-batch are
backpropagated directly; otherwise, we first assemble the complete score
matrix from detached scores and then replay the cell micro-batches using
gradients derived from the full family loss. This procedure preserves the
full family-level gradient while retaining activations for only one cell
micro-batch at a time. The implementation uses PyTorch~\citep{scholar_pytorch} 2.10.0, Transformers~\citep{scholar_transformers} 5.3.0, Accelerate 1.11.0, and DeepSpeed 0.18.4.

\clearpage
\section{Evaluation Setup}
\label{app:evaluation}

We re-run inference and evaluation for all compared methods under a common protocol, using public checkpoints when available, then public code, and finally our own reproduction. Appendix~\ref{app:compared-methods} lists the backbone and source for each method.

\subsection{Benchmark scoring}
  \label{app:eval}

\paragraph{Benchmarks and metrics.}
We evaluate fine-grained visual perception and hallucination on six benchmarks: HallusionBench~\citep{scholar_hallusionbench}, MMVP~\citep{scholar_mmvp}, AMBER~\citep{scholar_amber}, MMHal-Bench~\citep{scholar_mmhal}, CV-Bench-2D~\citep{scholar_cvbench}, and ObjectHal~\citep{scholar_chair}. We report aAcc and fAcc for HallusionBench, item and pair accuracy for MMVP, average accuracy for AMBER, hallucination rate for MMHal-Bench, macro accuracy for CV-Bench-2D, and CHAIRs for ObjectHal. Lower hallucination rate and CHAIRs indicate better performance; higher values are better for the other metrics.

  When a benchmark requires a model judge, we send a text-only request to
  \texttt{Qwen3.8-flash} with the temperature set to $0$. VLMEvalKit protocols follow the toolkit
  release~\citep{scholar_vlmevalkit}.
  \paragraph{MMVP.}
  Under VLMEvalKit's MMVP protocol, a binary hit
  $h_i\in\{0,1\}$ marks whether item $i$ matches the gold choice
  ($N=300$). Pair accuracy further requires both members of a consecutive
  pair to be right ($150$ pairs):
  \begin{equation*}
  \mathrm{ItemAcc}
    = \frac{1}{N}\sum_{i=1}^{N} h_i,
  \qquad
  \mathrm{PairAcc}
    = \frac{1}{N/2}\sum_{p=1}^{N/2} h_{2p-1}\,h_{2p}.
  \end{equation*}
  \paragraph{HallusionBench.}
  We evaluate all models on the same 951 questions with an image, grouped into 346 figures. We report answer-level accuracy (aAcc), the fraction of these questions answered correctly, and figure-level accuracy (fAcc), the fraction of figures for which all associated image-bearing questions are answered correctly.
  \paragraph{AMBER.}
  VLMEvalKit's AMBER rating maps replies to Yes/No
  (the judge is invoked only when the label is unresolved), then averages
  Attribute, Existence, and Relation accuracy with equal weight:
  \begin{equation*}
  \mathrm{AvgACC}
    = \frac{1}{3}
      \bigl(
        \mathrm{Acc}_{\mathrm{attr}}
        + \mathrm{Acc}_{\mathrm{exist}}
        + \mathrm{Acc}_{\mathrm{rel}}
      \bigr).
  \end{equation*}
  \paragraph{MMHal-Bench.}
  The five-shot MMHal rubric assigns each of the 96 answers a score from 0 to 6. We report the percentage of answers scoring below 3.
  \paragraph{CV-Bench-2D.}
  For the 1,438 questions, we compute accuracy separately for each source and report the unweighted mean of the source accuracies, following VLMEvalKit.
  \paragraph{ObjectHalBench.}
  The RLHF-V CHAIR pipeline matches object mentions in each caption against the COCO annotations for its image. We report $\mathrm{CHAIR}_s = 100\, n_{\mathrm{hall}}/300$, where $n_{\mathrm{hall}}$ is the number of captions containing at least one mention of an unannotated object.
\par\endgroup

\clearpage
\section{Qualitative Comparisons with the Base Model}
\label{app:qualitative-comparison}

\begin{figure}[H]
  \centering
  \includegraphics[page=1,width=\linewidth,height=0.80\textheight,keepaspectratio]{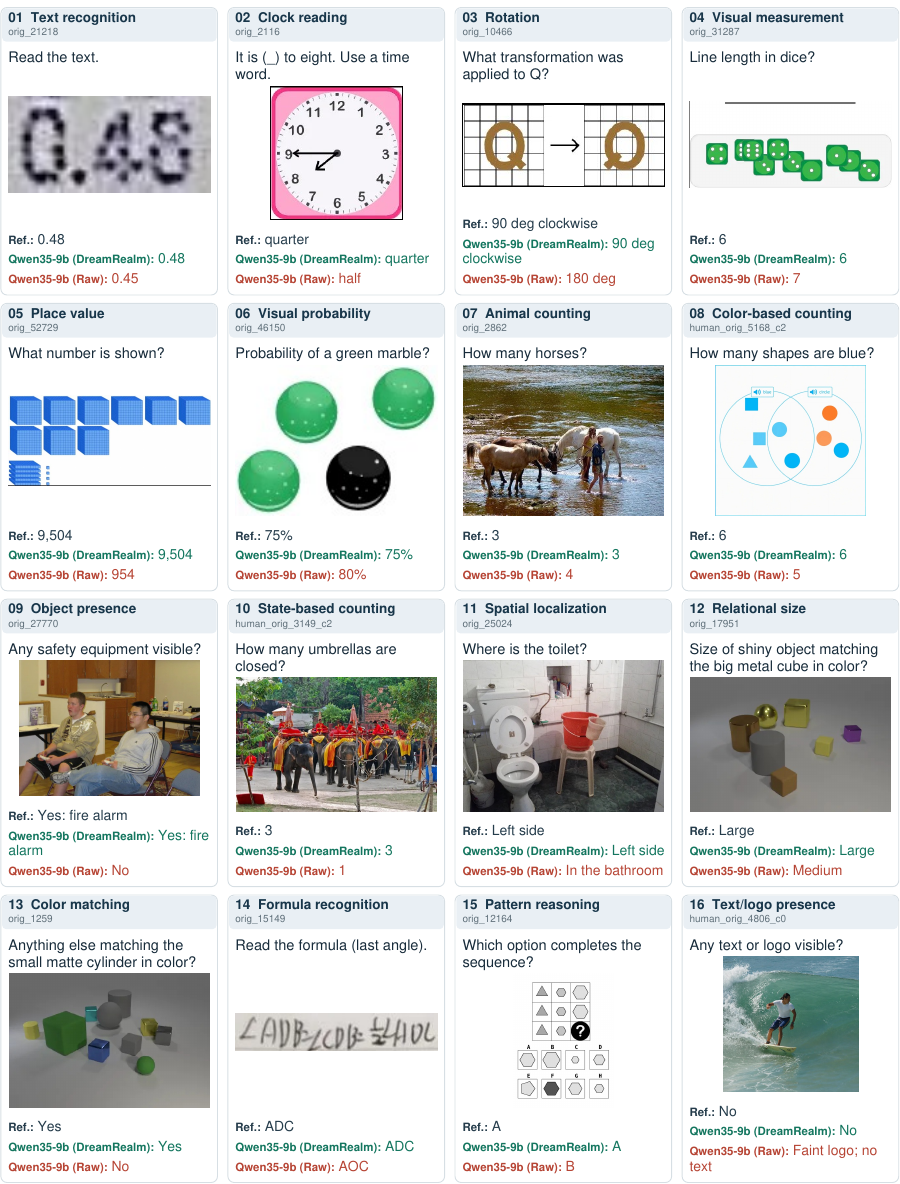}
  \caption{Compact qualitative comparisons (1/2), cases 1--16. Green and red identify model answers, not independent correctness judgments. Case 14 shows only the differing final angle in the formula. IDs beginning with \texttt{human\_} denote edited inputs.}
  \label{fig:showcase-1}
\end{figure}

\begin{figure}[H]
  \centering
  \includegraphics[page=2,width=\linewidth,height=0.92\textheight,keepaspectratio]{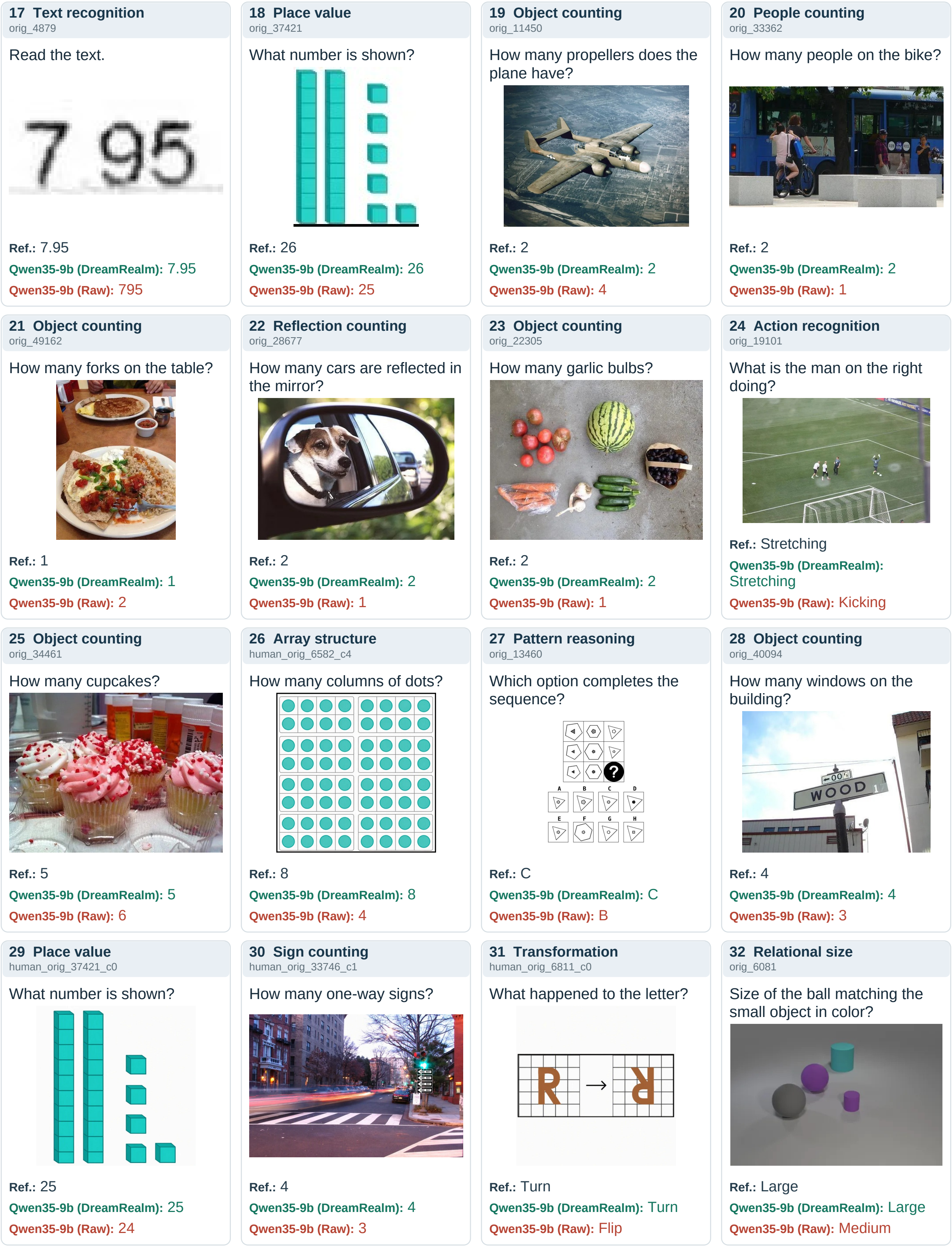}
  \caption{Compact qualitative comparisons (2/2), cases 17--32. Reflected vehicles in case 22 include the car carrying the dog. All responses are condensed from the recorded outputs.}
  \label{fig:showcase-2}
\end{figure}

Figures~\ref{fig:showcase-1}--\ref{fig:showcase-2} show 32 selected comparisons across diverse visual tasks. Each page contains 16 cases. Questions and answers are condensed while preserving their meaning; choice letters are expanded where useful. Ref. denotes the supplied reference; Qwen35-9b (DreamRealm) and Qwen35-9b (Raw) denote the adapted and original models, respectively. These selected examples do not estimate the overall success rate.

\clearpage

\clearpage

\section{Details of compared methods}
  \label{app:compared-methods}
  Block~I of Table~\ref{tab:main-comparison} reports specialized hallucination-mitigation
  methods. We evaluate these methods using the scoring rules in Appendix~\ref{app:eval}.
  For VCD and CoCI-CAP, we run inference-time decoding on the base model listed in Table~\ref{tab:block-i-backbones}. For VCD on MMHal-Bench, the contrastive image is an edited version of the input image, with the benchmark question used as the editing instruction. For the training-based methods, we use the public checkpoint when available and otherwise follow the published training recipe. The upper block therefore compares systems built on different base models.
  Within-backbone raw--adapted comparisons are in Block~II.
  \begin{table}[h]
  \centering
  \caption{Backbone and  setup for each Block~I method.}
  \label{tab:block-i-backbones}
  \small
  \begin{tabular}{@{}lll@{}}
  \toprule
  Method & Backbone & Setup \\
  \midrule
  VCD~\citep{scholar_vcd}
    & LLaVA-1.5-7B
    & Inference-time decoding \\
  CoCI-CAP~\citep{scholar_coci}
    & LLaVA-OneVision (Qwen2-7B)
    & Inference-time decoding \\
  RLAIF-V~\citep{scholar_rlaifv}
    & LLaVA-1.5-7B
    & Public checkpoint \\
  SymMPO~\citep{scholar_symmpo}
    & LLaVA-1.5-7B
    & Public checkpoint \\
  mDPO~\citep{scholar_mdpo}
    & Bunny (Llama-3-8B)
    & Reproduced  \\
  VC-STaR~\citep{scholar_vcstar}
    & Qwen2.5-VL-7B
    & Reproduced  \\
  MED~\citep{scholar_med}
    & Qwen2.5-VL-7B
    & Reproduced  \\
  IC-VCO~\citep{scholar_icvco}
    & LLaVA-OneVision (Qwen2-7B)
    & Reproduced  \\
  FINER~\citep{scholar_finer}
    & Qwen2.5-VL-7B
    & Public checkpoint \\
  IRIS~\citep{scholar_iris}
    & LLaVA-1.5-7B
    & Reproduced  \\
  C$^2$-DPO~\citep{scholar_c2dpo}
    & LLaVA-1.5-7B
    & Reproduced  \\
  \bottomrule
  \end{tabular}
  \end{table}

\section{Loss Component Ablation}
\label{app:loss-ablation}

\paragraph{Experimental setup.}
On Qwen3.5-9B, we compare the full RealmMatch objective with two variants that remove either the image contrast loss or the response contrast loss. Response contrast selects the correct response for a given image (Eq.~\ref{eq:realm-row}); image contrast selects the supporting image for a given response (Eq.~\ref{eq:realm-col}). Table~\ref{tab:loss-ablation} reports results alongside the original checkpoint (Raw).

\begin{table}[H]
\centering
\small
\caption{Loss ablations on Qwen3.5-9B. All values are percentages. Bold indicates the best result in each column. HallusionBench scores use only questions that include an image.}
\label{tab:loss-ablation}
\setlength{\tabcolsep}{3pt}
\begin{tabular*}{\linewidth}{@{\extracolsep{\fill}}lcccccccc@{}}
\toprule
& \multicolumn{2}{c}{HallusionBench} & \multicolumn{2}{c}{MMVP} & AMBER & MMHal & CV-2D & ObjectHal \\
\cmidrule(lr){2-3}\cmidrule(lr){4-5}
Configuration & aAcc$\uparrow$ & fAcc$\uparrow$ & Item$\uparrow$ & Pair$\uparrow$ & Avg.$\uparrow$ & Hall. rate$\downarrow$ & Macro$\uparrow$ & CHAIRs$\downarrow$ \\
\midrule
Raw & 71.61 & 54.91 & 80.00 & 63.33 & 88.34 & 47.92 & 82.16 & 30.33 \\
w/o $\mathcal{L}_{\mathrm{col}}$ & 76.55 & 53.47 & 84.67 & 71.33 & 89.30 & 32.29 & 82.41 & 10.67 \\
w/o $\mathcal{L}_{\mathrm{row}}$ & 75.18 & 53.76 & 86.33 & \textbf{75.33} & 90.82 & 37.50 & 82.60 & 11.00 \\
Full RealmMatch & \textbf{80.02} & \textbf{60.98} & \textbf{86.67} & 74.67 & \textbf{91.01} & \textbf{27.08} & \textbf{82.89} & \textbf{8.67} \\
\bottomrule
\end{tabular*}
\par\smallskip
\begin{minipage}{\linewidth}\footnotesize
\textit{Note.} Raw denotes the original Qwen3.5-9B model. The two ablation rows report results after removing the image contrast loss ($\mathcal{L}_{\mathrm{col}}$) or the response contrast loss ($\mathcal{L}_{\mathrm{row}}$) from RealmMatch, respectively.
\end{minipage}
\end{table}

Removing either contrastive loss reduces performance on at least seven of the eight metrics relative to full RealmMatch. Removing image contrast leads to larger drops than removing response contrast on five of the eight metrics.

\clearpage
\section{LLM Usage Statement}
RealmForgePipe mines hallucinations from each target model and uses GPT-Image-2 for image editing and Gemini 2.5 Pro for screening, edit planning, and verification (Appendices~\ref{app:materialization}, \ref{app:editing-walkthrough}, and~\ref{app:training-settings}). We used Qwen3.7-Max to assist with question-type labeling and Qwen3.8-Flash where benchmark protocols required an LLM judge (Appendices~\ref{app:data-categories} and~\ref{app:eval}).

We used OpenAI Codex to help implement experimental code, analyze benchmark results, and translate and edit the manuscript. We checked AI-assisted text against our methods, experimental results, and cited sources, revising or discarding suggestions as needed. We made all methodological and interpretive decisions and take full responsibility for the text, claims, data, and artifacts.

\section{Limitations}
\label{sec:limitations}
Applying DreamRealm to a new target model incurs costs for API calls to advanced models and for human review. These costs are relatively manageable in our setting. We report the cost breakdown in Appendices~\ref{app:data-construction} and~\ref{app:human-review}.

RealmSet inherits the uneven distribution of question types in VisCoR-55K, with counting questions accounting for a relatively large share. We did not rebalance these question types during data construction. Our experiments therefore do not fully explore DreamRealm's potential. Limited coverage of certain task types in the training data may help explain why gains on some benchmarks were modest.

Data construction in DreamRealm depends on the multimodal verifier's ability to diagnose errors and the image generation model's ability to carry out the resulting edits. If the verifier cannot diagnose an error even with access to the reference answer, or if the image generation model cannot carry out the proposed edit, the pipeline cannot produce a training example for that candidate.

\end{document}